\pdfoutput=1
\documentclass[11pt]{article}

\usepackage[preprint]{acl}

\usepackage{times}
\usepackage{latexsym}

\usepackage{amsmath}
\usepackage{diagbox}
\usepackage{enumitem}
\usepackage{subcaption}
\usepackage{graphicx}

\usepackage[T1]{fontenc}
\usepackage[utf8]{inputenc}

\usepackage{microtype}

\usepackage{inconsolata}

\title{Evaluating the fairness of task-adaptive pretraining on unlabeled test data before few-shot text classification}

\author{Kush Dubey \\
  Independent \\
  \texttt{kushdubey63@gmail.com}}

\begin{document}
\maketitle
\begin{abstract}
Few-shot learning benchmarks are critical for evaluating modern NLP techniques. It is possible, however, that benchmarks favor methods which easily make use of unlabeled text, because researchers can use unlabeled text from the test set to pretrain their models. Given the dearth of research on this potential problem, we run experiments to quantify the bias caused by pretraining on unlabeled test set text instead of on unlabeled, independently drawn text. Controlled few-shot and zero-shot experiments on 25 classification tasks and 3 language models---BERT, GPT-2, and Mistral 7B---do not find evidence of overoptimism. Furthermore, we demonstrate the importance of repeated subsampling when studying few-shot text classification, and recommend that few-shot learning benchmarks include multiple training folds. Code and data are available here: \url{https://github.com/kddubey/pretrain-on-test/}.
\end{abstract}

% TLDR: Experiments demonstrate that pretraining on unlabeled text from the test set does not bias test set performance as an estimator of out-of-sample performance.

\section{Introduction}
\label{sec:intro}

It is common for NLP benchmarks to release text from the test set, as researchers can submit a file of predictions instead of submitting code. A potential concern is that researchers can use this text during training. Consider the Real-world Annotated Few-shot Tasks (RAFT) benchmark \citep{alex2021raft}, which contains "few-shot" text classification tasks---tasks where the training set contains a relatively small number of labeled examples. Below is an excerpt from the RAFT paper (emphasis added):

\begin{quote}
    For each task, we release a public training set with 50 examples and a larger unlabeled test set. \textit{We encourage unsupervised pre-training on the unlabelled examples} and open-domain information retrieval.
\end{quote}

In the RAFT competition, a model is evaluated by scoring its predictions on the same set of unlabeled text which the model may have been trained on (using an unsupervised training procedure).

It is wrong to train a model on test set features with their labels and then evaluate on the test set when one needs to estimate performance on out-of-sample data. Test set performance would be overoptimistic \citep{hastie2009elements}. This fact is widely known. But what if, as encouraged by \citet{alex2021raft}, a model is trained on test set features \textit{without} test set labels? This paper studies this question for the domain of few-shot text classification.

\section{Motivation}

NLP benchmarks for few-shot learning are prevalent, as having only a handful of labeled examples is more realistic. One consideration when designing these benchmarks is that some few-shot approaches can---at least theoretically---use unlabeled text from the test set. With Pattern-Exploiting Training \citep{schick-schutze-2021-exploiting}, for example, one can train the final classifier on test set text with soft labels predicted by an ensemble of supervised models. With Pre-trained Prompt Tuning \citep{gu-etal-2022-ppt}, one can pretrain the language model (LM) on unlabeled test set text before prompt-tuning on the labeled training set. A more classical approach would be to train a word2vec model \citep{mikolov2013distributed} on unlabeled test set text, run this model on training text to get embeddings, and finally train a classifier on these embeddings with labels from the training set.

For other few-shot approaches, such as SetFit \citep{tunstall2022efficient} and in-context learning with LLMs (as popularized by \citealp{brown2020language}), it is more common to only use labeled text.

While the ability to exploit unlabeled text is useful, applying this ability to test set text could be substantively different than applying it to text which is statistically independent of the test set. This difference in methodology may be more concerning in the few-shot setting than in the many-shot setting. It is conceivable that differences between few-shot methods are due just as much to how unlabeled text is used as they are to how the few, labeled examples are used. This raises the question: does pretraining a model on a benchmark's unlabeled test set text inflate the model's performance on that benchmark?

\section{Related work}

As indicated by the quote in \S \ref{sec:intro}, the RAFT benchmark implicitly assumes that the answer is no. The validity of using test set features is not a fringe opinion. The popular textbook by \citet{hastie2009elements} contains the following passage without a reference or evidence (emphasis added):

\begin{quote}
    There is one qualification: \textit{initial unsupervised screening steps can be done before samples are left out.}
    For example, we could select the 1000 predictors with highest variance across all 50 samples, before starting cross-validation.
    \textit{Since this filtering does not involve the class labels, it does not give the predictors an unfair advantage.}
\end{quote}

The opposite opinion---that exploiting unlabeled test set features is unfair---may align more closely with best practices. For example, \citet{gururangan-etal-2020-dont} contains the following criticism of another study when comparing performances on a text classification task:

\begin{quote}
    \citet{thongtan-phienthrakul-2019-sentiment} report a higher number (97.42) on IMDB, but they
    train their word vectors on the test set.
\end{quote}

% 9-page
% \citet{jacovi-etal-2023-stop} argue that benchmarks which release unlabeled test set text can be compromised in two ways. First, the text can be discretely labeled and trained on by model developers. Second, if test set texts were scraped from websites which also host their labels, e.g., movie reviews and ratings on IMDb, then an LLM may have already been pretrained on these labeled texts (Scenario 1 in \citealp{jacovi-etal-2023-stop}). They do not discuss potential problems with using unlabeled test set text by itself.

% shorter version:
\citet{jacovi-etal-2023-stop} argue that benchmarks which release unlabeled test set text can be compromised, but do not discuss potential problems with using unlabeled test set text by itself.

\citet{moscovich2022cross} contains experiments and theory for unsupervised methods which are common to tasks involving tabular data.
% , namely: variance-based feature selection, grouping of rare categorical features, and feature rescaling.
They find that estimators of out-of-sample performance which were subject to these methods may be biased positively or negatively, depending on the parameters of the problem. They recommend further research on this bias in more domains, particularly when dealing with small sample sizes and high-dimensional data.

\section{Experimental design}
\label{sec:exp}

\begin{figure*}
\centering
\includegraphics[width=0.80\linewidth]{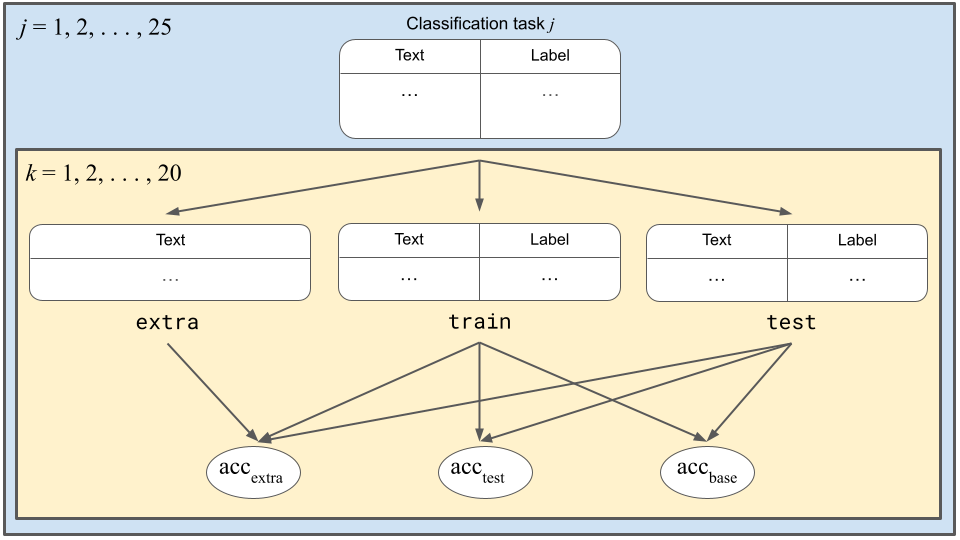} \hfill
\caption{The experimental design (\S \ref{sec:exp}) for $n = 500$ as an example.}
\label{fig:graph}
\end{figure*}

\begin{figure*}
\centering
\includegraphics[width=1.0\linewidth]{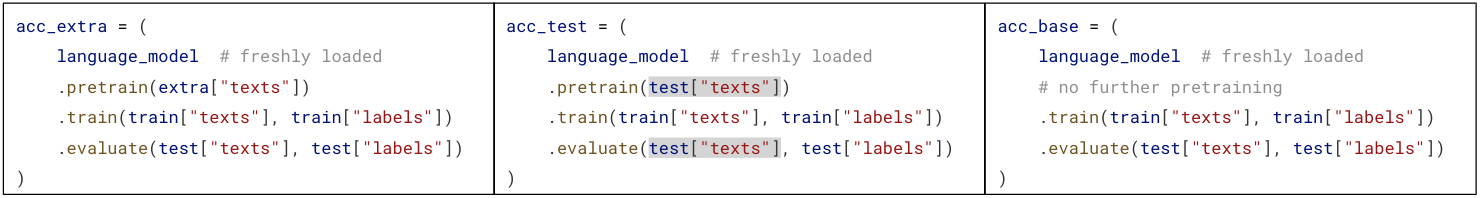} \hfill
\caption{Pseudocode for the accuracy estimators defined in \S \ref{sec:exp}.
% Notably, acc\textsubscript{test} is pretrained and evaluated on the same set of unlabeled text.
}
\label{fig:in_code}
\end{figure*}

We study whether pretraining on unlabeled test set text biases test set performance for 25 diverse text classification tasks and two types of LMs: BERT \citep{devlin-etal-2019-bert} and GPT-2 \citep{radford2019language}. Appendix~\ref{sec:tasks} describes each task.

The goal of the experiment is to first establish that pretraining is beneficial, in line with \citet{gururangan-etal-2020-dont}. Second, given that pretraining has a detectable effect, the experiment measures the accuracy difference between using test set text for the pretraining stage---an arguably unfair methodology---and using text which is independent of the test set---an inarguably fair methodology.

In more detail, the experiment starts by subsampling three separate sets of data from the full sample of data for a given text classification task:

\begin{itemize}[itemsep=0em]
    \item \texttt{extra}: $n$ (either $50, 100, 200$ or $500$) unlabeled texts which are optionally used for pretraining
    \item \texttt{train}: $m$ (either $50$ or $100$) labeled texts for classification training
    \item \texttt{test}: $n$ labeled texts to report accuracy.
\end{itemize}

\noindent Next, three accuracy estimators are computed. Procedures used to obtain them are described below.

\subsection{\texorpdfstring{acc\textsubscript{extra}}{acc extra}}
\label{sec:acc_extra}

\begin{enumerate}[itemsep=0em]
    \item Train a freshly loaded, pretrained LM on the $n$ unlabeled texts in \texttt{extra} using the LM's pretraining objective---masked language modeling loss for BERT, or causal language modeling loss for GPT-2. Texts are passed independently, and padded to form batches.
    \item Add a linear layer to this model and finetune all of the LM's weights to minimize classification cross entropy loss on \texttt{train}.
    % 9-page
    % Add in token embedding note from appendix
    \item Compute the classification accuracy of this model on \texttt{test}.
\end{enumerate}

Step 1 is task-adaptive pretraining---a procedure broadly recommended by \citet{gururangan-etal-2020-dont}. Step 2 is a canonical way to train a transformer-based LM for a classification task, according to Section 2 of \citet{zhang2021revisiting}.

acc\textsubscript{extra} is clearly an unbiased estimator of out-of-sample accuracy because it never trains on \texttt{test}. In other words, the expected value of acc\textsubscript{extra} is the accuracy one would observe on independent, identically distributed data.

\subsection{\texorpdfstring{acc\textsubscript{test}}{acc test}}

acc\textsubscript{test} is identical to acc\textsubscript{extra}, except that task-adaptive pretraining is done on unlabeled text from \texttt{test} instead of \texttt{extra} in step 1.

acc\textsubscript{test} represents what one might see in a competition like RAFT, where pretraining on unlabeled text from \texttt{test} is encouraged. It is unclear whether this accuracy estimator is unbiased, because it involved pretraining and evaluating on the same set of test set text. A reasonable hypothesis is that it is overoptimistic, i.e., $\text{E}[\text{acc\textsubscript{test}}] > \text{E}[\text{acc\textsubscript{extra}}]$.

\subsection{\texorpdfstring{acc\textsubscript{base}}{acc base}}
\label{sec:acc_base}

acc\textsubscript{base} does not do task-adaptive pretraining; it does not make any use of unlabeled text. It trains a pretrained LM on \texttt{train} to do classification, and then computes this model's accuracy on \texttt{test}. 

% \begin{enumerate}
%     \item Just finetune the whole model for classification on \texttt{train}.
%     \item Compute the classification accuracy of this model on \texttt{test}.
% \end{enumerate}

This score provides a sanity check. If there is no boost from acc\textsubscript{base} to acc\textsubscript{extra}, then it may not be surprising to observe no difference between acc\textsubscript{extra} and acc\textsubscript{test}. A boost from acc\textsubscript{base} to acc\textsubscript{extra} would rule out undertraining as the cause of a null difference between acc\textsubscript{extra} and acc\textsubscript{test} due to insufficient pretraining epochs or too low a learning rate.

\subsection{Repeated subsampling}
\label{sec:subsampling}

The accuracy estimators are paired, because their classification training and test data are identical. The only difference is the source of unlabeled text for pretraining. For acc\textsubscript{extra}, the source is independent of test data. For acc\textsubscript{test}, the test set text is used. For acc\textsubscript{base}, no unlabeled text is used.

A potentially important source of variation in this experiment is the particular subsamples, i.e., the particular realizations of \texttt{extra}, \texttt{train}, and \texttt{test} for a given classification task. To expose this variation, the experiment procedure is repeated tens of times for each task.\footnote{For $n = 50$ and $n = 100$, the experiment is repeated $100$ times. For $n = 200$, the experiment is repeated $50$ times. For $n = 500$, the experiment is repeated $20$ times. In total, $81,000$ finetuned BERT and GPT-2 models were evaluated.} For example, for $n = 500$, and for each of the $25$ tasks, $20$ (acc\textsubscript{extra}, acc\textsubscript{test}, acc\textsubscript{base}) triples are computed.

Appendix~\ref{sec:choices} explains more experiment choices.

\section{Results}
\label{sec:results}

Appendix~\ref{sec:diff_distr} visualizes the distributions of \textcolor{blue}{acc\textsubscript{extra} $-$ acc\textsubscript{base}} and \textcolor{orange}{acc\textsubscript{test} $-$ acc\textsubscript{extra}}. \textcolor{blue}{acc\textsubscript{extra} $-$ acc\textsubscript{base}} is a control: it is the accuracy boost from pretraining on unlabeled independent text versus not pretraining at all. \textcolor{orange}{acc\textsubscript{test} $-$ acc\textsubscript{extra}} is the main quantity of interest: it is the evaluation bias from pretraining on unlabeled test set text instead of on unlabeled independent text.

Table~\ref{tab:results} contains means of these differences for each configuration of the experiment. It roughly suggests that while pretraining is consistently beneficial, pretraining on unlabeled test set text does not bias test set performance one way or the other.

% A more complete analysis of this data is motivated and performed in the next section.

\begin{table}[ht]
    \centering
    \begin{subtable}[t]{0.45\textwidth}
        \centering
        \resizebox{\textwidth}{!}{
        \begin{tabular}{|l|l|l|}
        \hline
                & \multicolumn{1}{|c|}{BERT}                                             & \multicolumn{1}{|c|}{GPT-2}                                             \\ \hline
        $n=50$  & \diagbox[dir=SW]{\textcolor{blue}{4.1\%}}{\textcolor{orange}{0.19\%}}  & \diagbox[dir=SW]{\textcolor{blue}{3.8\%}}{\textcolor{orange}{0.18\%}}   \\ \hline
        $n=100$ & \diagbox[dir=SW]{\textcolor{blue}{3.9\%}}{\textcolor{orange}{0.18\%}}  & \diagbox[dir=SW]{\textcolor{blue}{4.1\%}}{\textcolor{orange}{0.11\%}}   \\ \hline
        $n=200$ & \diagbox[dir=SW]{\textcolor{blue}{3.9\%}}{\textcolor{orange}{-0.39\%}} & \diagbox[dir=SW]{\textcolor{blue}{4.4\%}}{\textcolor{orange}{-0.05\%}}  \\ \hline
        $n=500$ & \diagbox[dir=SW]{\textcolor{blue}{3.5\%}}{\textcolor{orange}{0.48\%}}  & \diagbox[dir=SW]{\textcolor{blue}{4.6\%}}{\textcolor{orange}{-0.08\%}}  \\ \hline
        \end{tabular}
        }
        \caption{$m = 50$}
        \label{tab:results_m50}
    \end{subtable}
    \hfill
    \begin{subtable}[t]{0.45\textwidth}
        \centering
        \resizebox{\textwidth}{!}{
        \begin{tabular}{|l|l|l|}
        \hline
                & \multicolumn{1}{|c|}{BERT}                                             & \multicolumn{1}{|c|}{GPT-2}                                             \\ \hline
        $n=50$  & \diagbox[dir=SW]{\textcolor{blue}{6.2\%}}{\textcolor{orange}{-0.08\%}} & \diagbox[dir=SW]{\textcolor{blue}{2.2\%}}{\textcolor{orange}{-0.05\%}}  \\ \hline
        $n=100$ & \diagbox[dir=SW]{\textcolor{blue}{6.1\%}}{\textcolor{orange}{-0.37\%}} & \diagbox[dir=SW]{\textcolor{blue}{2.5\%}}{\textcolor{orange}{0.03\%}}   \\ \hline
        $n=200$ & \diagbox[dir=SW]{\textcolor{blue}{4.1\%}}{\textcolor{orange}{0.33\%}}  & \diagbox[dir=SW]{\textcolor{blue}{6.3\%}}{\textcolor{orange}{-0.01\%}}  \\ \hline
        $n=500$ & \diagbox[dir=SW]{\textcolor{blue}{6.1\%}}{\textcolor{orange}{-0.16\%}} & \diagbox[dir=SW]{\textcolor{blue}{3.9\%}}{\textcolor{orange}{-0.21\%}}  \\ \hline
        \end{tabular}
        }
        \caption{$m = 100$}
        \label{tab:results_m100}
    \end{subtable}
    \caption{Means of accuracy differences taken across all subsamples of all 25 classification tasks. For each cell, the upper-left of the diagonal corresponds to the sample mean of \textcolor{blue}{acc\textsubscript{extra} $-$ acc\textsubscript{base}}, and the lower-right corresponds to the sample mean of \textcolor{orange}{acc\textsubscript{test} $-$ acc\textsubscript{extra}}.}
    \label{tab:results}
\end{table}

\section{Analysis}
\label{sec:analysis}

Reporting means is not enough, especially when studying few-shot learning. Appendix~\ref{sec:diff_distr} demonstrates that there is considerable variance, despite pairing the accuracy estimators.\footnote{One source of variance is intentionally introduced: the subsample splits, as explained in \S \ref{sec:subsampling}. The other source of variance is inherent: the added linear layer to perform classification is initialized with random weights.} While these visualizations tell us about how raw accuracy differences vary, they do not tell us how the mean accuracy difference varies. We seek a neat answer to the core questions: on this benchmark of 25 classification tasks, how much does the overall accuracy differ between two modeling techniques, and how much does this difference vary?

One way to communicate the variance is to estimate the standard error of the mean difference across classification tasks. But the standard error statistic can be difficult to interpret \citep{morey2016fallacy}. Furthermore, its computation is not completely trivial due to the data's hierarchical dependency structure: each triple, (acc\textsubscript{extra}, acc\textsubscript{test}, acc\textsubscript{base}), is drawn from (\texttt{train}, \texttt{test}), which is itself drawn from the given classification dataset.

\begin{figure*}[t]
\includegraphics[width=0.48\linewidth]{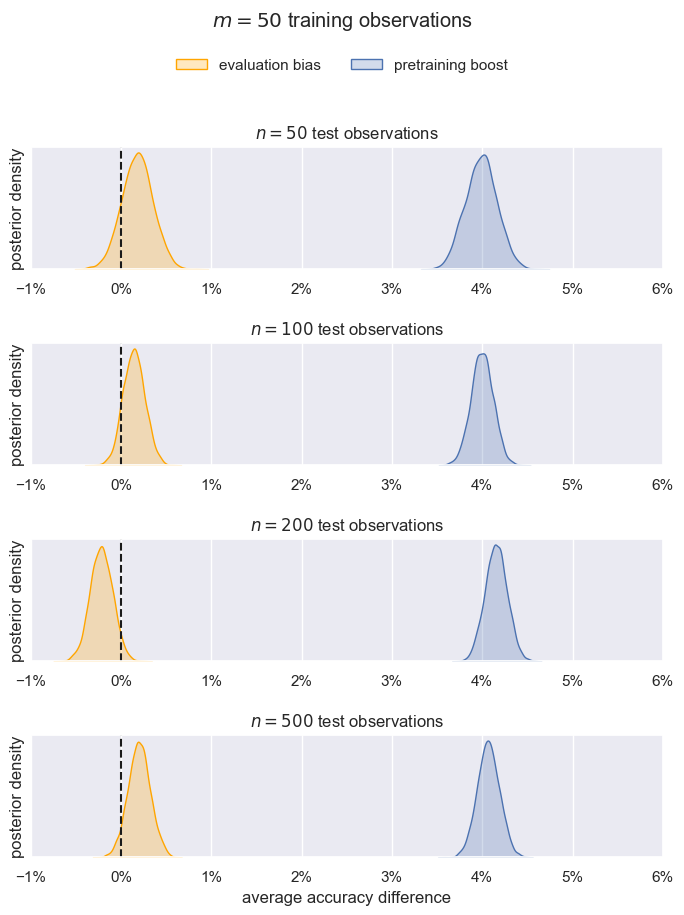} \hfill
\includegraphics[width=0.48\linewidth]{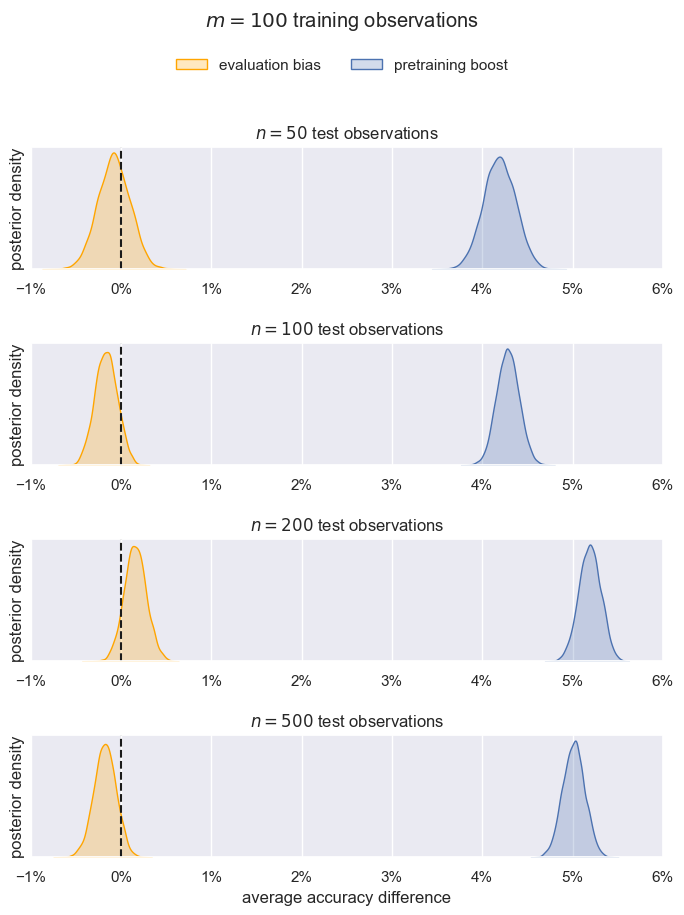}
\caption{Distributions of average accuracy differences \eqref{eq:marg}. The evaluation bias is akin to \textcolor{orange}{acc\textsubscript{test} $-$ acc\textsubscript{extra}}. The pretraining boost is akin to \textcolor{blue}{acc\textsubscript{extra} $-$ acc\textsubscript{base}}.}
\label{fig:main}
\end{figure*}

\subsection{Model}
\label{sec:model}

This analysis does not aim to estimate standard errors. Instead, a hierarchical model is fit. Specifically, for each LM type (indexed by $i = 1, 2$ for BERT and GPT-2), each classification task (indexed by $j = 1, 2, \dots, 25$), each of their subsamples (indexed by $k = 1, 2, \dots, 20$ for $n = 500$, for example), and a control and treatment (indexed by $l = 0, 1$), the number of correct predictions is modeled ($*$ is short for $ijkl$):

\begin{align}
Y_{*} \sim \text{Binomial}(n, \lambda_{*}) \\
\resizebox{0.896\columnwidth}{!}{$\text{logit}(\lambda_{*}) = \mu + \alpha z_i + U_j + V_{jk} + W_{jl} + \beta x_{l}$} \\
\mu \sim \text{Normal}(0, 1) \\
\alpha \sim \text{Normal}(0, 5) \\
U_j \sim \text{Normal}(0, \sigma_{U}) \\
V_{jk} \sim \text{Normal}(0, \sigma_{V}) \\
W_{jl} \sim \text{Normal}(0, \sigma_{W}) \\
\beta \sim \text{Normal}(0, 1) \\
\sigma_{U}, \sigma_{V} \sim \text{HalfNormal}(0, 1) \\
&\hspace*{-4.85cm} \sigma_{W} \sim \text{HalfNormal}(0, 3.5355)
\end{align}

\begin{enumerate}[label={(\arabic*)},itemsep=0em]
    \item number of correct predictions
    \item logit link for accuracy rate, additive effects
    \item prior for the global intercept
    \item prior for the effect of the type of LM (BERT or GPT-2)---a control variable
    \item prior for the effect of the classification task (partial-pooled to reduce overfitting)
    \item prior for the nested effect of the task's subsampled dataset
    \item prior for the interaction effect of the task and the intervention (to reduce underfitting)
    \item prior for the effect of the intervention
    \item prior for standard deviations
    \item prior for standard deviation.
\end{enumerate}

% In-line, shorter version:
% \begin{align*}
% Y_{*} \sim \text{Binomial}(n, \lambda_{*}) && \text{number of correct predictions} \\
% \text{logit}(\lambda_{*}) = \mu + \alpha z_i + U_j + V_{jk} + W_{jl} + \beta x_{l} && \text{logit link accuracy rate, additive effects} \\
% \mu \sim \text{Normal}(0, 1) && \text{prior for the global intercept} \\
% \alpha \sim \text{Normal}(0, 5) && \text{prior for LM effect} \\
% U_j \sim \text{Normal}(0, \sigma_{U}) && \text{prior for task effect} \\
% V_{jk} \sim \text{Normal}(0, \sigma_{V}) && \text{prior for nested effect of subsample} \\
% W_{jl} \sim \text{Normal}(0, \sigma_{W}) && \text{prior for interaction of task, intervention} \\
% \beta \sim \text{Normal}(0, 1) && \text{prior for intervention} \\
% \sigma_{U}, \sigma_{V} \sim \text{HalfNormal}(0, 1) && \text{prior for standard deviations} \\
% \sigma_{W} \sim \text{HalfNormal}(0, 3.5355) && \text{prior for standard deviation.}
% \end{align*}

The model is fit using Markov Chain Monte Carlo, using the interface provided by the \texttt{bambi} package \citep{Capretto2022}.

To analyze the pretraining boost, the control, $Y_{ijk0}$, is $n \cdot$ acc\textsubscript{base}, and the treatment, $Y_{ijk1}$, is $n \cdot$ acc\textsubscript{extra}. Here, the intervention refers to pretraining on unlabeled independent text versus not pretraining at all.

To analyze the evaluation bias, the control, $Y_{ijk0}$, is $n \cdot$ acc\textsubscript{extra}, and the treatment, $Y_{ijk1}$, is $n \cdot$ acc\textsubscript{test}. Here, the intervention refers to pretraining on unlabeled text from the test set instead of on unlabeled independent text.

4,000 samples from the posterior predictive, $\hat{Y}_{ijkl}$, are drawn. Appendix~\ref{sec:model_checks} includes a simulation demonstrating the model's ability to correctly recover null and non-null effects.

\subsection{Overall effects}
\label{sec:overall_effects}

Benchmarks assess methods by taking their average performance across tasks. To place the results in this context, samples from the posterior predictive distribution of $Y_{ijk1} - Y_{ijk0}$ (\ref{sec:model}) are taken, then averaged across $i$ (the $2$ LM types---BERT and GPT-2), $j$ (the $25$ classification tasks), and $k$ (their subsamples), and divided by $n$ to obtain the distribution of the average accuracy difference (expressed in dot notation, where dots are used as placeholders for indices that have been averaged over):

\begin{align}
\frac{\bar{\hat{Y}}_{\cdot \cdot \cdot 1} - \bar{\hat{Y}}_{\cdot \cdot \cdot 0}}{n}. \label{eq:marg}
\end{align}

Each distribution is that of the marginal effect of the modeling intervention: pretraining versus not pretraining (the pretraining boost), or pretraining on unlabeled test set text instead of on unlabeled independent text (the evaluation bias).

\subsection{Task-level effects}
\label{sec:task_level_effects}

While taking an average across tasks provides a concise summary, it cannot be used to rule out the existence of an evaluation bias. If the direction of the bias depends on latent properties of the task, averaging may cancel out real, positive biases with real, negative ones. Alternatively, it may dilute the few real, positive biases with many null ones.

\citet{jin-etal-2021-causal} argue and demonstrate that the benefit of task-adaptive pretraining depends on the task's causal direction. If the principle of independent causal mechanisms is also relevant to the fairness of pretraining on test set features, then our accuracy data may contain (for the sake of argument) positive evaluation biases for anti-causal tasks, and null biases for causal tasks.\footnote{We will not assess any particular hypothesis about the role of causality. We are only motivating task-level analysis.}

One way to analyze tasks is to sample from the posterior predictive distribution of the accuracy difference, and only average across subsamples:

\begin{align}
\frac{\bar{\hat{Y}}_{i j \cdot 1} - \bar{\hat{Y}}_{i j \cdot 0}}{n}. \label{eq:cond}
\end{align}

A more concise way is to perform a hypothesis test for each setting of $m, n$, and the LM type:

\begin{align}
H_0&: \text{E}[\textcolor{orange}{\text{acc}_{\text{test}} - \text{acc}_{\text{extra}}}] = 0 \label{eq:null} \\
H_1&: \text{E}[\textcolor{orange}{\text{acc}_{\text{test}} - \text{acc}_{\text{extra}}}] > 0.
\end{align}

The $p$-value is estimated via permutation testing. It is then adjusted to control the false discovery rate \citep{benjamini1995controlling}.

\begin{figure*}[!htb]
\includegraphics[width=0.48\linewidth]{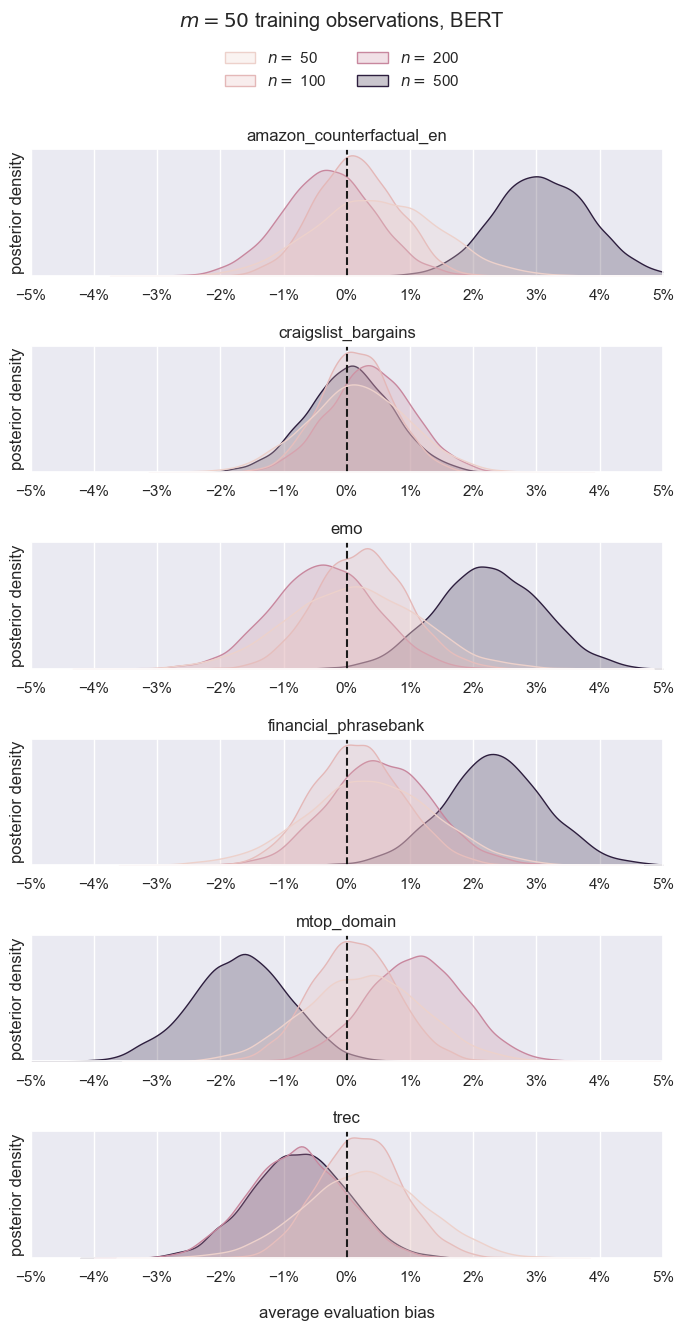} \hfill
\includegraphics[width=0.48\linewidth]{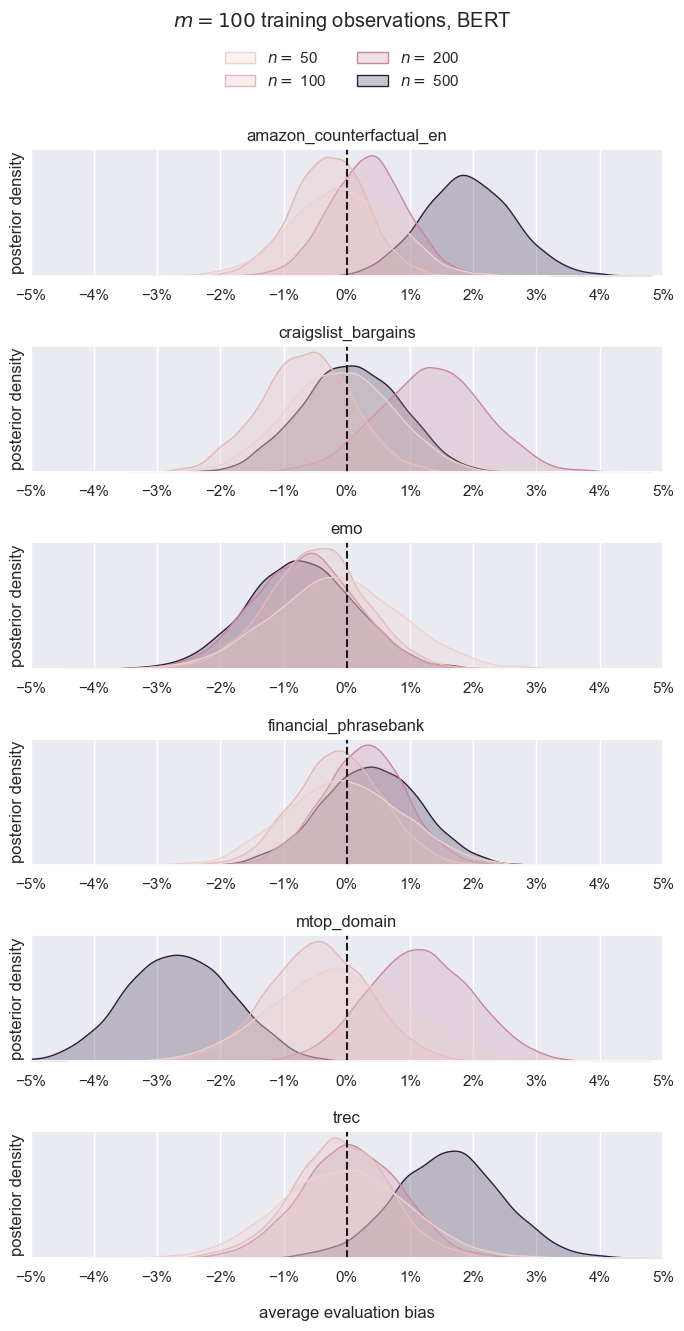}
\caption{Distributions of average evaluation biases \eqref{eq:cond} for the subset of tasks which reported an average evaluation bias of at least +3\% accuracy in any configuration of the experiment.
% Note that for some tasks, such as \texttt{clickbait\_notclickbait\_dataset}, the posterior predictive distribution shrank the raw evaluation bias from +3\% more towards 0 than others.
}
\label{fig:lemons_bert}
\end{figure*}

\section{Discussion}
\label{sec:discussion}

Figure~\ref{fig:main} demonstrates that the average pretraining boost is significant in every configuration of the experiment. This finding replicates that from \citet{gururangan-etal-2020-dont}. After averaging across settings for $m$, $n$, and the 2 LM types, only two of the 25 classification tasks had a pretraining boost less than 0, and both were greater than -1\%.
% 9-page
\footnote{The tasks were \href{https://huggingface.co/datasets/blog_authorship_corpus}{\texttt{blog\_authorship\_corpus}} and \href{https://huggingface.co/datasets/movie_rationales}{\texttt{movie\_rationales}}.}
Task-adaptive pretraining had the intended effect.

As shown in Figure~\ref{fig:main}, the evaluation bias bounces inconsistently and insignificantly around 0. After averaging, 12 of the 25 classification tasks had a positive evaluation bias, 13 had a negative evaluation bias, and all tasks had an average evaluation bias less than 1\% in absolute value.

To avoid excessive averaging, we lemon-picked tasks which reported a bias of at least +3\% in any experiment configuration. All tasks matching this criterion were from experiments with BERT, as BERT had greater training variance. If there were a task-dependent evaluation bias, one could expect that the bias is consistent across $m$ or $n$ within a task, or there is a consistent pattern with how the bias changes with $m$ or $n$ across tasks. Figure~\ref{fig:lemons_bert} does not clearly support either of these hypotheses.

% 9-page
% TODO: put this in the first paragraph of conclusion?
% Given the lack of evidence for an evaluation bias, it is unlikely that a benchmark which releases unlabeled test set text systematically promotes models pretrained on it over equally performant models which pretrained on unlabeled independent text.

\citet{moscovich2022cross} found that the evaluation bias caused by unsupervised methods for tabular data converges to $0$ as $n$ increases. This finding is not confirmed by this experiment. Figure~\ref{fig:main} shows that within $m=50$ and $m=100$, distributions of the evaluation bias hover around 0 across $n$. Figure~\ref{fig:lemons_bert} also does not support a relationship between $n$ and the evaluation bias for lemon-picked tasks. But far more experiments varying $n$ are needed to thoroughly assess this insensitivity.

\section{Overtraining}
\label{sec:overtraining}
 
\S \ref{sec:discussion} rules out undertraining on unlabeled text as the cause of a null evaluation bias. What if we overtrain? Overtraining on labeled test data trivially increases test set performance. Perhaps overtraining on unlabeled test set text has a similar effect. To test this hypothesis for text classification, GPT-2 is intentionally overtrained on unlabeled text for 2 epochs instead of 1.
% pretraining can only exploit noise in the unlabeled text which the model is then rewarded for when it is evaluated on that same text.

\begin{figure*}
\includegraphics[width=0.48\linewidth]{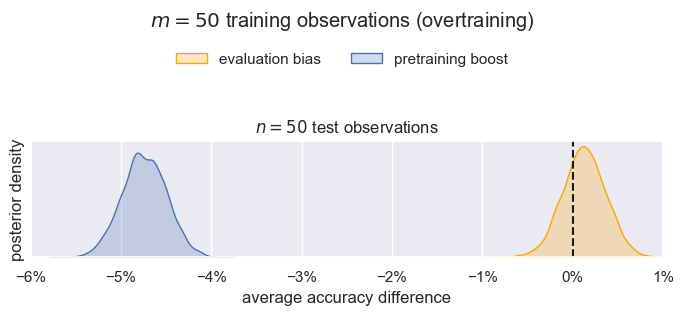} \hfill
\includegraphics[width=0.48\linewidth]{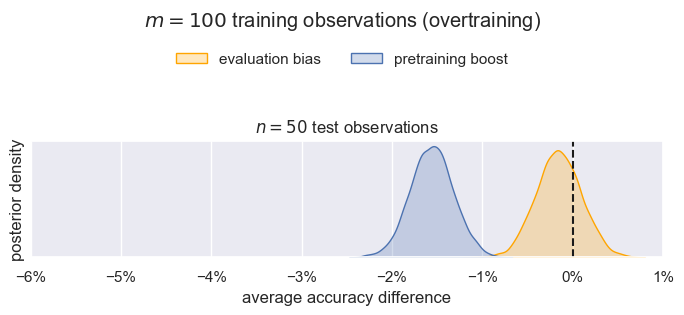}
\caption{Average accuracy differences \eqref{eq:marg} after pretraining GPT-2 for 2 epochs instead of 1 (\S \ref{sec:overtraining}).}
\label{fig:overtraining}
\end{figure*}

\begin{figure*}
\includegraphics[width=0.48\linewidth]{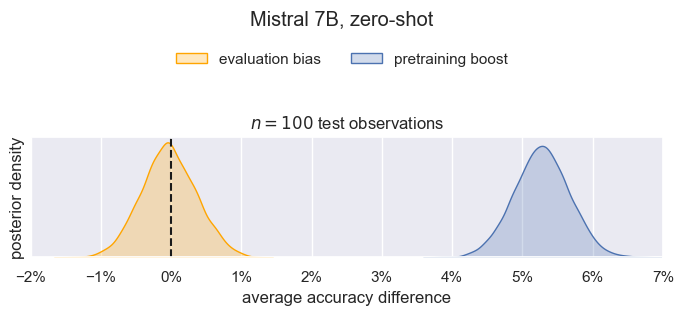} \hfill
\includegraphics[width=0.48\linewidth]{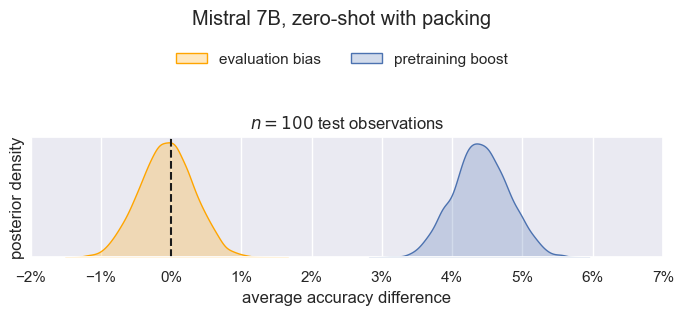}
\caption{Average accuracy differences \eqref{eq:marg} for zero-shot classification (\S \ref{sec:zero_shot}) with padding (left) and packing (right). For each of the 25 classification tasks, 20 subsamples were taken.}
\label{fig:zero_shot}
\end{figure*}

For each of the 25 classification tasks and their subsamples, pretraining for 2 epochs instead of 1 resulted in a lower pretraining loss. The final pretraining loss is 20\% lower on average, and the pretraining boost is negative, which indicates overfitting, as intended. Figure~\ref{fig:overtraining} demonstrates that, despite overtraining, the evaluation bias hovers around 0. All 50 $p$-values from the test in \eqref{eq:null} are greater than $0.5$.\footnote{Note that all $p$-values from the test in \eqref{eq:null} are adjusted to control the false discovery rate.} Overtraining on unlabeled test set text causes test set performance to degrade to the same degree that overfitting on unlabeled independent text does.

\section{Zero-shot text classification}
\label{sec:zero_shot}

Prompting an LLM is a popular choice for solving NLP problems. These prompts can be pretrained on. For example, Gemma 2 \citep{team2024gemma} is intentionally pretrained on prompts from the LMSYS benchmark \citep{zheng2023lmsys}.

To study a more modern prompting approach, the experiment in \S \ref{sec:exp} is repeated with two modifications. First, task-adaptive pretraining is done on prompts---unlabeled texts with instructions for solving the task. Second, classification training is not performed; \texttt{train} is unused. The further-pretrained LLM is immediately prompted to do the task on \texttt{test}.

More specifically, pretraining is performed by adding a QLoRA adapter layer \citep{dettmers2024qlora} to every linear layer in Mistral 7B \citep{jiang2023mistral}. Perhaps notably, instructions mention the set of possible answers---the class names.

% 9-page
% Here is an example of a prompt for the \href{https://huggingface.co/datasets/fancyzhx/ag_news}{\texttt{ag\_news}} task \citep{zhang2015character}:

Figure~\ref{fig:zero_shot} (left) shows that, while pretraining on prompts improves accuracy, pretraining on test set prompts does not increase test set accuracy compared to pretraining on independently drawn prompts. 12 of the 25 tasks had a positive evaluation bias and 13 had a negative evaluation bias. All 25 $p$-values from \eqref{eq:null} are greater than $0.5$; there is no evidence of a task-level evaluation bias.

A limitation of this experiment is that it does not account for contamination. If Mistral 7B's pretraining data included labeled or unlabeled parts of the datasets used here, the pretraining boost and evaluation bias may be diluted.

\subsection{Packing instead of padding}
\label{sec:packing}

Experiments so far passed pretraining texts independently, adding and masking pad tokens to enable batching. Packing instead combines texts into a single sequence of tokens whose length is the model's context length. Packing is often used during the initial pretraining of an LLM, where the model is trained on continuous streams of text to increase throughput \citep{brown2020language}.

Does packing impact evaluation bias differently than padding? One hypothesis is that, without special handling of the attention mask,
packing causes the model to attend to previous texts, so the transformer has greater flexibility in modeling unlabeled text. To study the effects of packing, the zero-shot experiment in \S \ref{sec:zero_shot} is repeated with packing instead of padding. Figure~\ref{fig:zero_shot} (right) shows that there is a pretraining boost, but no evaluation bias. All 25 $p$-values from \eqref{eq:null} are greater than $0.5$.

\subsection{On testing test set contamination}
\label{sec:contamination}

Contamination detectors aim to flag overoptimistic LLM evaluations. An LLM is contaminated if it was pretrained and evaluated on the same set of labeled data, as this procedure results in an evaluation bias. In contrast, the result from \S \ref{sec:packing} implies that contamination of \textit{unlabeled} test set text does not result in an evaluation bias. Do contamination detectors pick up this nuance?

The experiment in \S \ref{sec:packing} is run for the \texttt{ag\_news} task and $n = 500$. Next, text-label pairs from \texttt{test} are passed to the contamination hypothesis test in \citet{DBLP:conf/iclr/OrenMCLH24}.
% 9-page
% This test computes the $p$-value by comparing the average log-probability of shuffled texts with sequences in the order they were packed.
% As is, this test does not distinguish between labeled and unlabeled text.
The $p$-value for the model pretrained on unlabeled text from \texttt{extra} is 0.33. The $p$-value for the model pretrained on unlabeled text from \texttt{test} is 0.015, which indicates contamination.
% 9-page
% \footnote{A caveat of this experiment is that the $p$-value for tasks with long answers is likely higher.}
However, the observed evaluation bias for this task is statistically indistinguishable from 0.

Detectors need to be able to differentiate the contamination of labeled text from the contamination of unlabeled text. For those that do not, contamination flags should be interpreted with care. Even if such a detector never raises false flags, a contamination flag may not indicate an overoptimistic evaluation.

\begin{figure*}[ht!]
\includegraphics[width=0.48\linewidth]{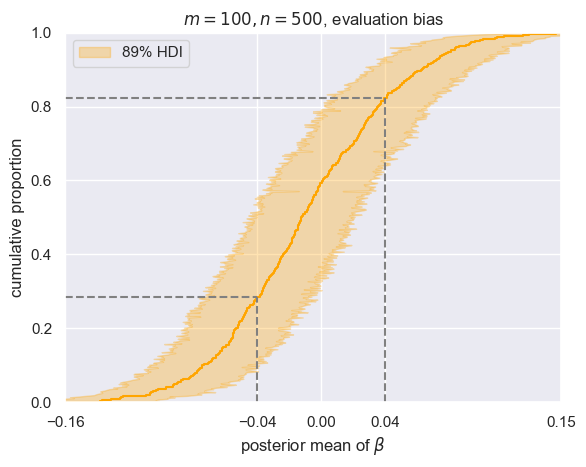} \hfill
\includegraphics[width=0.48\linewidth]{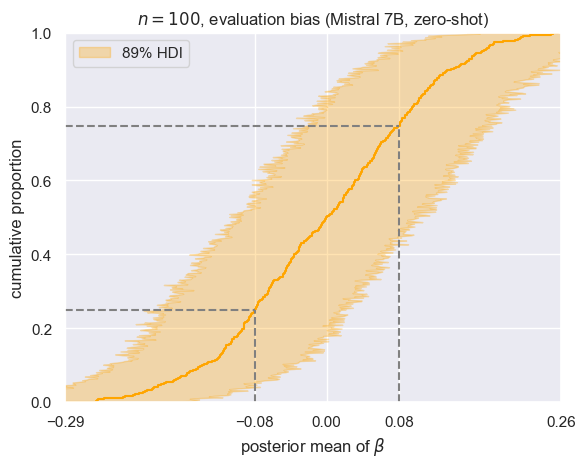}
\caption{Distributions of conclusions had there been no technical replication (\S \ref{sec:meta}).}
\label{fig:meta}
\end{figure*}

\section{Meta-analysis}
\label{sec:meta}

\S \ref{sec:subsampling} briefly argues for subsampling multiple datasets from the full classification dataset. To assess this argument, the analysis was repeated on $500$ random slices of the $m=100, n=500$ dataset of accuracies such that exactly $1$ (acc\textsubscript{extra}, acc\textsubscript{test}, acc\textsubscript{base}) triple per classification task (instead of $20$ triples) is included. This de-replicated data is often all one gets from benchmarks.

Figure~\ref{fig:meta} (left) displays the cumulative distribution of the posterior mean of the evaluation bias for $m=100, n=500$ under this de-replicated experimental design. The distribution is quite variant. There is a 47\% chance that the posterior mean of $\beta$---the average increase in the log-odds of a correct prediction by pretraining on unlabeled test set text instead of on unlabeled independent text---is outside the interval $(-0.04, 0.04)$, which would indicate a significant negative or positive bias.\footnote{For $0.04$, the odds ratio is $e^{0.04} \approx 1.04$. For context, the average odds ratio between adjacent submissions in the \href{https://huggingface.co/spaces/ought/raft-leaderboard}{RAFT leaderboard} is $1.03$. For posterior means outside $(-0.04, 0.04)$, all of their 89\% credible intervals exclude 0, which evidences a non-null effect.} For the zero-shot experiment in \S \ref{sec:zero_shot}, there is a 50\% chance that that the posterior mean of $\beta$ is outside $(-0.08, 0.08)$. Without repeated subsampling, one may as well flip a coin to decide whether pretraining on unlabeled test set text is fair.

\section{Conclusion}

Task-adaptive pretraining on unlabeled test set text---instead of on unlabeled independent text---did not result in a consistent or significant evaluation bias. This appears to be the case when pretraining helps, when it hurts, and when pretraining is done on texts with instructions.

For benchmarks which release unlabeled text from the test set, this finding does not completely absolve LLM evaluations from scrutiny. The reason is that the boost from pretraining on unlabeled text---which is often significant---could be viewed as a type of evaluation bias, depending on how LLMs generalize. More concretely, suppose there is a benchmark and two LLMs, $A$ and $B$. $A$ was \textit{not} pretrained on the benchmark's unlabeled test set text, while $B$ was. With the perspective that LLM benchmarks supply scores which are correlates of performance on real-world tasks---instead of indicators of performance solely on the benchmark's tasks---then $B$ scoring higher on the benchmark than $A$ may be a misleading signal. If pretraining on the benchmark's unlabeled text causes $B$ to generalize better only \textit{within} the distribution of the benchmark, then $B$'s edge on this benchmark does not signal an edge in real-world tasks. Knowing whether an LLM was pretrained on unlabeled test set text is still important.

One recommendation for designing few-shot benchmarks, which expands on the principle about robustness from \citet{bragg2021flex}, is based on the meta-analysis in \S \ref{sec:meta}: empirical studies of few-shot learning should consider including multiple, independent subsamples of training data. While a single training set combined with a large test set is sufficient for precise, unbiased estimation of out-of-sample performance, this estimator is conditional on the training set. In few-shot learning, the training set is, by definition, minimal. The estimator hides two sources of variance---that from the randomly drawn training set, and that from randomness inherent in the training procedure. Figure~\ref{fig:meta} shows that this variance is large-enough to turn a methodology into a coin flip for two different training procedures. In-context learning with LLMs is also sensitive to the selection of few-shot examples (\citealp{lu-etal-2022-fantastically}, \citealp{alzahrani2024benchmarks}). Benchmarks which require training on multiple, independent subsamples would expose training variance.

\section*{Limitations}

This paper does not study semi-supervised methods like Pattern-Exploiting Training, or hand-inspecting the test set text and targeting interventions accordingly. We also do not study the effect of including unlabeled test set texts in the initial pretraining stage of an LLM.

The results are empirical. There may be tasks where an evaluation bias exists, and these were not part of the 25 classification tasks we collected. The results do not theoretically or universally establish that pretraining on unlabeled test set text is fair.

\section*{Acknowledgements}

The author is grateful to Eilon Reisin-Tzur for valuable feedback and insightful discussions, and for motivating the question addressed in Appendix~\ref{sec:meta-appendix}. The author is also grateful to his family for the continued provision of sustenance, shelter, and crossword puzzles.

\bibliography{anthology,custom}

\begin{thebibliography}{52}
\expandafter\ifx\csname natexlab\endcsname\relax\def\natexlab#1{#1}\fi

\bibitem[{Alex et~al.(2021)Alex, Lifland, Tunstall, Thakur, Maham, Riedel, Hine, Ashurst, Sedille, Carlier, Noetel, and Stuhlm\"{u}ller}]{alex2021raft}
Neel Alex, Eli Lifland, Lewis Tunstall, Abhishek Thakur, Pegah Maham, C.~Riedel, Emmie Hine, Carolyn Ashurst, Paul Sedille, Alexis Carlier, Michael Noetel, and Andreas Stuhlm\"{u}ller. 2021.
\newblock \href {https://arxiv.org/abs/2109.14076} {Raft: A real-world few-shot text classification benchmark}.
\newblock In \emph{Proceedings of the Neural Information Processing Systems Track on Datasets and Benchmarks}, volume~1.

\bibitem[{Alzahrani et~al.(2024)Alzahrani, Alyahya, Alnumay, Alrashed, Alsubaie, Almushaykeh, Mirza, Alotaibi, Altwairesh, Alowisheq et~al.}]{alzahrani2024benchmarks}
Norah Alzahrani, Hisham~Abdullah Alyahya, Yazeed Alnumay, Sultan Alrashed, Shaykhah Alsubaie, Yusef Almushaykeh, Faisal Mirza, Nouf Alotaibi, Nora Altwairesh, Areeb Alowisheq, et~al. 2024.
\newblock \href {https://arxiv.org/abs/2402.01781} {When benchmarks are targets: Revealing the sensitivity of large language model leaderboards}.
\newblock \emph{arXiv preprint arXiv:2402.01781}.

\bibitem[{Benjamini and Hochberg(1995)}]{benjamini1995controlling}
Yoav Benjamini and Yosef Hochberg. 1995.
\newblock \href {https://doi.org/https://doi.org/10.1111/j.2517-6161.1995.tb02031.x} {Controlling the false discovery rate: a practical and powerful approach to multiple testing}.
\newblock \emph{Journal of the Royal statistical society: series B (Methodological)}, 57(1):289--300.

\bibitem[{Bragg et~al.(2021)Bragg, Cohan, Lo, and Beltagy}]{bragg2021flex}
Jonathan Bragg, Arman Cohan, Kyle Lo, and Iz~Beltagy. 2021.
\newblock \href {https://arxiv.org/abs/2107.07170} {Flex: Unifying evaluation for few-shot nlp}.
\newblock \emph{Advances in Neural Information Processing Systems}, 34:15787--15800.

\bibitem[{Brown et~al.(2020)Brown, Mann, Ryder, Subbiah, Kaplan, Dhariwal, Neelakantan, Shyam, Sastry, Askell et~al.}]{brown2020language}
Tom Brown, Benjamin Mann, Nick Ryder, Melanie Subbiah, Jared~D Kaplan, Prafulla Dhariwal, Arvind Neelakantan, Pranav Shyam, Girish Sastry, Amanda Askell, et~al. 2020.
\newblock \href {https://arxiv.org/abs/2005.14165} {Language models are few-shot learners}.
\newblock \emph{Advances in neural information processing systems}, 33:1877--1901.

\bibitem[{Capretto et~al.(2022)Capretto, Piho, Kumar, Westfall, Yarkoni, and Martin}]{Capretto2022}
Tomás Capretto, Camen Piho, Ravin Kumar, Jacob Westfall, Tal Yarkoni, and Osvaldo~A Martin. 2022.
\newblock \href {https://doi.org/10.18637/jss.v103.i15} {Bambi: A simple interface for fitting bayesian linear models in python}.
\newblock \emph{Journal of Statistical Software}, 103(15):1–29.

\bibitem[{Chapuis et~al.(2020)Chapuis, Colombo, Manica, Labeau, and Clavel}]{chapuis-etal-2020-hierarchical}
Emile Chapuis, Pierre Colombo, Matteo Manica, Matthieu Labeau, and Chlo{\'e} Clavel. 2020.
\newblock \href {https://doi.org/10.18653/v1/2020.findings-emnlp.239} {Hierarchical pre-training for sequence labelling in spoken dialog}.
\newblock In \emph{Findings of the Association for Computational Linguistics: EMNLP 2020}, pages 2636--2648, Online. Association for Computational Linguistics.

\bibitem[{Chatterjee et~al.(2019)Chatterjee, Narahari, Joshi, and Agrawal}]{chatterjee-etal-2019-semeval}
Ankush Chatterjee, Kedhar~Nath Narahari, Meghana Joshi, and Puneet Agrawal. 2019.
\newblock \href {https://doi.org/10.18653/v1/S19-2005} {{S}em{E}val-2019 task 3: {E}mo{C}ontext contextual emotion detection in text}.
\newblock In \emph{Proceedings of the 13th International Workshop on Semantic Evaluation}, pages 39--48, Minneapolis, Minnesota, USA. Association for Computational Linguistics.

\bibitem[{Dettmers et~al.(2024)Dettmers, Pagnoni, Holtzman, and Zettlemoyer}]{dettmers2024qlora}
Tim Dettmers, Artidoro Pagnoni, Ari Holtzman, and Luke Zettlemoyer. 2024.
\newblock \href {https://arxiv.org/abs/2305.14314} {Qlora: Efficient finetuning of quantized llms}.
\newblock \emph{Advances in Neural Information Processing Systems}, 36.

\bibitem[{Devlin et~al.(2019)Devlin, Chang, Lee, and Toutanova}]{devlin-etal-2019-bert}
Jacob Devlin, Ming-Wei Chang, Kenton Lee, and Kristina Toutanova. 2019.
\newblock \href {https://doi.org/10.18653/v1/N19-1423} {{BERT}: Pre-training of deep bidirectional transformers for language understanding}.
\newblock In \emph{Proceedings of the 2019 Conference of the North {A}merican Chapter of the Association for Computational Linguistics: Human Language Technologies, Volume 1 (Long and Short Papers)}, pages 4171--4186, Minneapolis, Minnesota. Association for Computational Linguistics.

\bibitem[{DeYoung et~al.(2020)DeYoung, Jain, Rajani, Lehman, Xiong, Socher, and Wallace}]{deyoung-etal-2020-eraser}
Jay DeYoung, Sarthak Jain, Nazneen~Fatema Rajani, Eric Lehman, Caiming Xiong, Richard Socher, and Byron~C. Wallace. 2020.
\newblock \href {https://doi.org/10.18653/v1/2020.acl-main.408} {{ERASER}: {A} benchmark to evaluate rationalized {NLP} models}.
\newblock In \emph{Proceedings of the 58th Annual Meeting of the Association for Computational Linguistics}, pages 4443--4458, Online. Association for Computational Linguistics.

\bibitem[{Diggelmann et~al.(2020)Diggelmann, Boyd-Graber, Bulian, Ciaramita, and Leippold}]{diggelmann2020climatefever}
Thomas Diggelmann, Jordan Boyd-Graber, Jannis Bulian, Massimiliano Ciaramita, and Markus Leippold. 2020.
\newblock \href {http://arxiv.org/abs/2012.00614} {Climate-fever: A dataset for verification of real-world climate claims}.

\bibitem[{FitzGerald et~al.(2023)FitzGerald, Hench, Peris, Mackie, Rottmann, Sanchez, Nash, Urbach, Kakarala, Singh, Ranganath, Crist, Britan, Leeuwis, Tur, and Natarajan}]{fitzgerald-etal-2023-massive}
Jack FitzGerald, Christopher Hench, Charith Peris, Scott Mackie, Kay Rottmann, Ana Sanchez, Aaron Nash, Liam Urbach, Vishesh Kakarala, Richa Singh, Swetha Ranganath, Laurie Crist, Misha Britan, Wouter Leeuwis, Gokhan Tur, and Prem Natarajan. 2023.
\newblock \href {https://doi.org/10.18653/v1/2023.acl-long.235} {{MASSIVE}: A 1{M}-example multilingual natural language understanding dataset with 51 typologically-diverse languages}.
\newblock In \emph{Proceedings of the 61st Annual Meeting of the Association for Computational Linguistics (Volume 1: Long Papers)}, pages 4277--4302, Toronto, Canada. Association for Computational Linguistics.

\bibitem[{Grano et~al.(2017)Grano, Di~Sorbo, Mercaldo, Visaggio, Canfora, and Panichella}]{grano2017android}
Giovanni Grano, Andrea Di~Sorbo, Francesco Mercaldo, Corrado~A Visaggio, Gerardo Canfora, and Sebastiano Panichella. 2017.
\newblock \href {https://dl.acm.org/doi/10.1145/3121264.3121266} {Android apps and user feedback: a dataset for software evolution and quality improvement}.
\newblock In \emph{Proceedings of the 2nd ACM SIGSOFT international workshop on app market analytics}, pages 8--11.

\bibitem[{Gu et~al.(2022)Gu, Han, Liu, and Huang}]{gu-etal-2022-ppt}
Yuxian Gu, Xu~Han, Zhiyuan Liu, and Minlie Huang. 2022.
\newblock \href {https://doi.org/10.18653/v1/2022.acl-long.576} {{PPT}: Pre-trained prompt tuning for few-shot learning}.
\newblock In \emph{Proceedings of the 60th Annual Meeting of the Association for Computational Linguistics (Volume 1: Long Papers)}, pages 8410--8423, Dublin, Ireland. Association for Computational Linguistics.

\bibitem[{Guha et~al.(2024)Guha, Nyarko, Ho, R{\'e}, Chilton, Chohlas-Wood, Peters, Waldon, Rockmore, Zambrano et~al.}]{guha2024legalbench}
Neel Guha, Julian Nyarko, Daniel Ho, Christopher R{\'e}, Adam Chilton, Alex Chohlas-Wood, Austin Peters, Brandon Waldon, Daniel Rockmore, Diego Zambrano, et~al. 2024.
\newblock \href {https://arxiv.org/abs/2308.11462} {Legalbench: A collaboratively built benchmark for measuring legal reasoning in large language models}.
\newblock \emph{Advances in Neural Information Processing Systems}, 36.

\bibitem[{Gururangan et~al.(2020)Gururangan, Marasovi{\'c}, Swayamdipta, Lo, Beltagy, Downey, and Smith}]{gururangan-etal-2020-dont}
Suchin Gururangan, Ana Marasovi{\'c}, Swabha Swayamdipta, Kyle Lo, Iz~Beltagy, Doug Downey, and Noah~A. Smith. 2020.
\newblock \href {https://doi.org/10.18653/v1/2020.acl-main.740} {Don{'}t stop pretraining: Adapt language models to domains and tasks}.
\newblock In \emph{Proceedings of the 58th Annual Meeting of the Association for Computational Linguistics}, pages 8342--8360, Online. Association for Computational Linguistics.

\bibitem[{Hastie et~al.(2009)Hastie, Tibshirani, Friedman, and Friedman}]{hastie2009elements}
Trevor Hastie, Robert Tibshirani, Jerome~H Friedman, and Jerome~H Friedman. 2009.
\newblock \href {https://hastie.su.domains/Papers/ESLII.pdf} {\emph{The elements of statistical learning: data mining, inference, and prediction}}, volume~2.
\newblock Springer.

\bibitem[{He et~al.(2018)He, Chen, Balakrishnan, and Liang}]{he-etal-2018-decoupling}
He~He, Derek Chen, Anusha Balakrishnan, and Percy Liang. 2018.
\newblock \href {https://doi.org/10.18653/v1/D18-1256} {Decoupling strategy and generation in negotiation dialogues}.
\newblock In \emph{Proceedings of the 2018 Conference on Empirical Methods in Natural Language Processing}, pages 2333--2343, Brussels, Belgium. Association for Computational Linguistics.

\bibitem[{Huangzhao(2018)}]{zhang2018}
Zhang Huangzhao. 2018.
\newblock Yahoo-answers-topic-classification-dataset.
\newblock \url{https://github.com/LC-John/Yahoo-Answers-Topic-Classification-Dataset}.

\bibitem[{Jacovi et~al.(2023)Jacovi, Caciularu, Goldman, and Goldberg}]{jacovi-etal-2023-stop}
Alon Jacovi, Avi Caciularu, Omer Goldman, and Yoav Goldberg. 2023.
\newblock \href {https://doi.org/10.18653/v1/2023.emnlp-main.308} {Stop uploading test data in plain text: Practical strategies for mitigating data contamination by evaluation benchmarks}.
\newblock In \emph{Proceedings of the 2023 Conference on Empirical Methods in Natural Language Processing}, pages 5075--5084, Singapore. Association for Computational Linguistics.

\bibitem[{Jiang et~al.(2023)Jiang, Sablayrolles, Mensch, Bamford, Chaplot, Casas, Bressand, Lengyel, Lample, Saulnier et~al.}]{jiang2023mistral}
Albert~Q Jiang, Alexandre Sablayrolles, Arthur Mensch, Chris Bamford, Devendra~Singh Chaplot, Diego de~las Casas, Florian Bressand, Gianna Lengyel, Guillaume Lample, Lucile Saulnier, et~al. 2023.
\newblock \href {https://arxiv.org/abs/2310.06825} {Mistral 7b}.
\newblock \emph{arXiv preprint arXiv:2310.06825}.

\bibitem[{Jin et~al.(2021)Jin, von K{\"u}gelgen, Ni, Vaidhya, Kaushal, Sachan, and Schoelkopf}]{jin-etal-2021-causal}
Zhijing Jin, Julius von K{\"u}gelgen, Jingwei Ni, Tejas Vaidhya, Ayush Kaushal, Mrinmaya Sachan, and Bernhard Schoelkopf. 2021.
\newblock \href {https://doi.org/10.18653/v1/2021.emnlp-main.748} {Causal direction of data collection matters: Implications of causal and anticausal learning for {NLP}}.
\newblock In \emph{Proceedings of the 2021 Conference on Empirical Methods in Natural Language Processing}, pages 9499--9513, Online and Punta Cana, Dominican Republic. Association for Computational Linguistics.

\bibitem[{Kiesel et~al.(2019)Kiesel, Mestre, Shukla, Vincent, Adineh, Corney, Stein, and Potthast}]{kiesel-etal-2019-semeval}
Johannes Kiesel, Maria Mestre, Rishabh Shukla, Emmanuel Vincent, Payam Adineh, David Corney, Benno Stein, and Martin Potthast. 2019.
\newblock \href {https://doi.org/10.18653/v1/S19-2145} {{S}em{E}val-2019 task 4: Hyperpartisan news detection}.
\newblock In \emph{Proceedings of the 13th International Workshop on Semantic Evaluation}, pages 829--839, Minneapolis, Minnesota, USA. Association for Computational Linguistics.

\bibitem[{Kumar et~al.(2019)Kumar, Carroll, Hartikainen, and Martin}]{arviz_2019}
Ravin Kumar, Colin Carroll, Ari Hartikainen, and Osvaldo Martin. 2019.
\newblock \href {https://doi.org/10.21105/joss.01143} {Arviz a unified library for exploratory analysis of bayesian models in python}.
\newblock \emph{Journal of Open Source Software}, 4(33):1143.

\bibitem[{Ljubešić et~al.(2019)Ljubešić, Fišer, and Erjavec}]{ljubešić2019frenk}
Nikola Ljubešić, Darja Fišer, and Tomaž Erjavec. 2019.
\newblock \href {http://arxiv.org/abs/1906.02045} {The frenk datasets of socially unacceptable discourse in slovene and english}.

\bibitem[{Lu et~al.(2022)Lu, Bartolo, Moore, Riedel, and Stenetorp}]{lu-etal-2022-fantastically}
Yao Lu, Max Bartolo, Alastair Moore, Sebastian Riedel, and Pontus Stenetorp. 2022.
\newblock \href {https://doi.org/10.18653/v1/2022.acl-long.556} {Fantastically ordered prompts and where to find them: Overcoming few-shot prompt order sensitivity}.
\newblock In \emph{Proceedings of the 60th Annual Meeting of the Association for Computational Linguistics (Volume 1: Long Papers)}, pages 8086--8098, Dublin, Ireland. Association for Computational Linguistics.

\bibitem[{Malo et~al.(2014)Malo, Sinha, Korhonen, Wallenius, and Takala}]{Malo2014GoodDO}
P.~Malo, A.~Sinha, P.~Korhonen, J.~Wallenius, and P.~Takala. 2014.
\newblock \href {https://arxiv.org/abs/1307.5336} {Good debt or bad debt: Detecting semantic orientations in economic texts}.
\newblock \emph{Journal of the Association for Information Science and Technology}, 65.

\bibitem[{Manotas et~al.(2020)Manotas, Vo, and Sheinin}]{manotas-etal-2020-limit}
Irene Manotas, Ngoc Phuoc~An Vo, and Vadim Sheinin. 2020.
\newblock \href {https://doi.org/10.18653/v1/2020.findings-emnlp.88} {{L}i{M}i{T}: The literal motion in text dataset}.
\newblock In \emph{Findings of the Association for Computational Linguistics: EMNLP 2020}, pages 991--1000, Online. Association for Computational Linguistics.

\bibitem[{McElreath(2018)}]{mcelreath2018statistical}
Richard McElreath. 2018.
\newblock \href {https://xcelab.net/rm/statistical-rethinking/} {\emph{Statistical rethinking: A Bayesian course with examples in R and Stan}}.
\newblock Chapman and Hall/CRC.

\bibitem[{Metsis et~al.(2006)Metsis, Androutsopoulos, and Paliouras}]{metsis2006spam}
Vangelis Metsis, Ion Androutsopoulos, and Georgios Paliouras. 2006.
\newblock \href {https://www2.aueb.gr/users/ion/docs/ceas2006_paper.pdf} {Spam filtering with naive bayes-which naive bayes?}
\newblock In \emph{CEAS}, volume~17, pages 28--69. Mountain View, CA.

\bibitem[{Mikolov et~al.(2013)Mikolov, Sutskever, Chen, Corrado, and Dean}]{mikolov2013distributed}
Tomas Mikolov, Ilya Sutskever, Kai Chen, Greg~S Corrado, and Jeff Dean. 2013.
\newblock \href {https://arxiv.org/abs/1310.4546} {Distributed representations of words and phrases and their compositionality}.
\newblock \emph{Advances in neural information processing systems}, 26.

\bibitem[{Morey et~al.(2016)Morey, Hoekstra, Rouder, Lee, and Wagenmakers}]{morey2016fallacy}
Richard~D Morey, Rink Hoekstra, Jeffrey~N Rouder, Michael~D Lee, and Eric-Jan Wagenmakers. 2016.
\newblock \href {https://pubmed.ncbi.nlm.nih.gov/26450628/} {The fallacy of placing confidence in confidence intervals}.
\newblock \emph{Psychonomic bulletin \& review}, 23:103--123.

\bibitem[{Moscovich and Rosset(2022)}]{moscovich2022cross}
Amit Moscovich and Saharon Rosset. 2022.
\newblock \href {https://arxiv.org/abs/1901.08974} {On the cross-validation bias due to unsupervised preprocessing}.
\newblock \emph{Journal of the Royal Statistical Society Series B: Statistical Methodology}, 84(4):1474--1502.

\bibitem[{Muennighoff et~al.(2023)Muennighoff, Tazi, Magne, and Reimers}]{muennighoff-etal-2023-mteb}
Niklas Muennighoff, Nouamane Tazi, Loic Magne, and Nils Reimers. 2023.
\newblock \href {https://doi.org/10.18653/v1/2023.eacl-main.148} {{MTEB}: Massive text embedding benchmark}.
\newblock In \emph{Proceedings of the 17th Conference of the European Chapter of the Association for Computational Linguistics}, pages 2014--2037, Dubrovnik, Croatia. Association for Computational Linguistics.

\bibitem[{O{'}Neill et~al.(2021)O{'}Neill, Rozenshtein, Kiryo, Kubota, and Bollegala}]{oneill-etal-2021-wish}
James O{'}Neill, Polina Rozenshtein, Ryuichi Kiryo, Motoko Kubota, and Danushka Bollegala. 2021.
\newblock \href {https://doi.org/10.18653/v1/2021.emnlp-main.568} {{I} wish {I} would have loved this one, but {I} didn{'}t {--} a multilingual dataset for counterfactual detection in product review}.
\newblock In \emph{Proceedings of the 2021 Conference on Empirical Methods in Natural Language Processing}, pages 7092--7108, Online and Punta Cana, Dominican Republic. Association for Computational Linguistics.

\bibitem[{Oren et~al.(2024)Oren, Meister, Chatterji, Ladhak, and Hashimoto}]{DBLP:conf/iclr/OrenMCLH24}
Yonatan Oren, Nicole Meister, Niladri~S. Chatterji, Faisal Ladhak, and Tatsunori Hashimoto. 2024.
\newblock \href {https://openreview.net/forum?id=KS8mIvetg2} {Proving test set contamination in black-box language models}.
\newblock In \emph{The Twelfth International Conference on Learning Representations, {ICLR} 2024, Vienna, Austria, May 7-11, 2024}. OpenReview.net.

\bibitem[{Pang and Lee(2005)}]{pang-lee-2005-seeing}
Bo~Pang and Lillian Lee. 2005.
\newblock \href {https://doi.org/10.3115/1219840.1219855} {Seeing stars: Exploiting class relationships for sentiment categorization with respect to rating scales}.
\newblock In \emph{Proceedings of the 43rd Annual Meeting of the Association for Computational Linguistics ({ACL}{'}05)}, pages 115--124, Ann Arbor, Michigan. Association for Computational Linguistics.

\bibitem[{Radford et~al.(2019)Radford, Wu, Child, Luan, Amodei, Sutskever et~al.}]{radford2019language}
Alec Radford, Jeffrey Wu, Rewon Child, David Luan, Dario Amodei, Ilya Sutskever, et~al. 2019.
\newblock \href {https://paperswithcode.com/paper/language-models-are-unsupervised-multitask} {Language models are unsupervised multitask learners}.
\newblock \emph{OpenAI blog}, 1(8):9.

\bibitem[{Saravia et~al.(2018)Saravia, Liu, Huang, Wu, and Chen}]{saravia-etal-2018-carer}
Elvis Saravia, Hsien-Chi~Toby Liu, Yen-Hao Huang, Junlin Wu, and Yi-Shin Chen. 2018.
\newblock \href {https://doi.org/10.18653/v1/D18-1404} {{CARER}: Contextualized affect representations for emotion recognition}.
\newblock In \emph{Proceedings of the 2018 Conference on Empirical Methods in Natural Language Processing}, pages 3687--3697, Brussels, Belgium. Association for Computational Linguistics.

\bibitem[{Schick and Sch{\"u}tze(2021)}]{schick-schutze-2021-exploiting}
Timo Schick and Hinrich Sch{\"u}tze. 2021.
\newblock \href {https://doi.org/10.18653/v1/2021.eacl-main.20} {Exploiting cloze-questions for few-shot text classification and natural language inference}.
\newblock In \emph{Proceedings of the 16th Conference of the European Chapter of the Association for Computational Linguistics: Main Volume}, pages 255--269, Online. Association for Computational Linguistics.

\bibitem[{Schler et~al.(2006)Schler, Koppel, Argamon, and Pennebaker}]{schler2006effects}
Jonathan Schler, Moshe Koppel, Shlomo Argamon, and James~W Pennebaker. 2006.
\newblock \href {https://cdn.aaai.org/Symposia/Spring/2006/SS-06-03/SS06-03-039.pdf} {Effects of age and gender on blogging.}
\newblock In \emph{AAAI spring symposium: Computational approaches to analyzing weblogs}, volume~6, pages 199--205.

\bibitem[{Sharma et~al.(2019)Sharma, Li, and Wang}]{sharma-etal-2019-bigpatent}
Eva Sharma, Chen Li, and Lu~Wang. 2019.
\newblock \href {https://doi.org/10.18653/v1/P19-1212} {{BIGPATENT}: A large-scale dataset for abstractive and coherent summarization}.
\newblock In \emph{Proceedings of the 57th Annual Meeting of the Association for Computational Linguistics}, pages 2204--2213, Florence, Italy. Association for Computational Linguistics.

\bibitem[{Sharma(2019)}]{sharma2019}
Roshan Sharma. 2019.
\newblock Twitter-sentiment-analysis.
\newblock \url{https://github.com/sharmaroshan/Twitter-Sentiment-Analysis}.

\bibitem[{Team et~al.(2024)Team, Riviere, Pathak, Sessa, Hardin, Bhupatiraju, Hussenot, Mesnard, Shahriari, Ram{\'e} et~al.}]{team2024gemma}
Gemma Team, Morgane Riviere, Shreya Pathak, Pier~Giuseppe Sessa, Cassidy Hardin, Surya Bhupatiraju, L{\'e}onard Hussenot, Thomas Mesnard, Bobak Shahriari, Alexandre Ram{\'e}, et~al. 2024.
\newblock \href {https://arxiv.org/abs/2408.00118} {Gemma 2: Improving open language models at a practical size}.
\newblock \emph{arXiv preprint arXiv:2408.00118}.

\bibitem[{Thongtan and Phienthrakul(2019)}]{thongtan-phienthrakul-2019-sentiment}
Tan Thongtan and Tanasanee Phienthrakul. 2019.
\newblock \href {https://doi.org/10.18653/v1/P19-2057} {Sentiment classification using document embeddings trained with cosine similarity}.
\newblock In \emph{Proceedings of the 57th Annual Meeting of the Association for Computational Linguistics: Student Research Workshop}, pages 407--414, Florence, Italy. Association for Computational Linguistics.

\bibitem[{Tunstall et~al.(2022)Tunstall, Reimers, Jo, Bates, Korat, Wasserblat, and Pereg}]{tunstall2022efficient}
Lewis Tunstall, Nils Reimers, Unso Eun~Seo Jo, Luke Bates, Daniel Korat, Moshe Wasserblat, and Oren Pereg. 2022.
\newblock \href {https://arxiv.org/abs/2209.11055} {Efficient few-shot learning without prompts}.
\newblock \emph{arXiv preprint arXiv:2209.11055}.

\bibitem[{Wang et~al.(2007)Wang, Smith, and Mitamura}]{wang-etal-2007-jeopardy}
Mengqiu Wang, Noah~A. Smith, and Teruko Mitamura. 2007.
\newblock \href {https://aclanthology.org/D07-1003} {What is the {J}eopardy model? a quasi-synchronous grammar for {QA}}.
\newblock In \emph{Proceedings of the 2007 Joint Conference on Empirical Methods in Natural Language Processing and Computational Natural Language Learning ({EMNLP}-{C}o{NLL})}, pages 22--32, Prague, Czech Republic. Association for Computational Linguistics.

\bibitem[{Wolf et~al.(2020)Wolf, Debut, Sanh, Chaumond, Delangue, Moi, Cistac, Rault, Louf, Funtowicz, Davison, Shleifer, von Platen, Ma, Jernite, Plu, Xu, Le~Scao, Gugger, Drame, Lhoest, and Rush}]{wolf-etal-2020-transformers}
Thomas Wolf, Lysandre Debut, Victor Sanh, Julien Chaumond, Clement Delangue, Anthony Moi, Pierric Cistac, Tim Rault, Remi Louf, Morgan Funtowicz, Joe Davison, Sam Shleifer, Patrick von Platen, Clara Ma, Yacine Jernite, Julien Plu, Canwen Xu, Teven Le~Scao, Sylvain Gugger, Mariama Drame, Quentin Lhoest, and Alexander Rush. 2020.
\newblock \href {https://doi.org/10.18653/v1/2020.emnlp-demos.6} {Transformers: State-of-the-art natural language processing}.
\newblock In \emph{Proceedings of the 2020 Conference on Empirical Methods in Natural Language Processing: System Demonstrations}, pages 38--45, Online. Association for Computational Linguistics.

\bibitem[{Zhang et~al.(2021)Zhang, Wu, Katiyar, Weinberger, and Artzi}]{zhang2021revisiting}
Tianyi Zhang, Felix Wu, Arzoo Katiyar, Kilian~Q Weinberger, and Yoav Artzi. 2021.
\newblock \href {https://arxiv.org/abs/2006.05987} {Revisiting few-sample bert fine-tuning}.
\newblock In \emph{International Conference on Learning Representations}.

\bibitem[{Zhang et~al.(2015)Zhang, Zhao, and LeCun}]{zhang2015character}
Xiang Zhang, Junbo Zhao, and Yann LeCun. 2015.
\newblock \href {https://arxiv.org/abs/1509.01626} {Character-level convolutional networks for text classification}.
\newblock \emph{Advances in neural information processing systems}, 28.

\bibitem[{Zheng et~al.(2023)Zheng, Chiang, Sheng, Li, Zhuang, Wu, Zhuang, Li, Lin, Xing et~al.}]{zheng2023lmsys}
Lianmin Zheng, Wei-Lin Chiang, Ying Sheng, Tianle Li, Siyuan Zhuang, Zhanghao Wu, Yonghao Zhuang, Zhuohan Li, Zi~Lin, Eric Xing, et~al. 2023.
\newblock \href {https://arxiv.org/abs/2309.11998} {Lmsys-chat-1m: A large-scale real-world llm conversation dataset}.
\newblock \emph{arXiv preprint arXiv:2309.11998}.

\end{thebibliography}

\appendix

\section{Classification tasks}
\label{sec:tasks}

\begin{table*}
\begin{tabular}{|l|l|l|l|}
\hline
\textbf{Hugging Face dataset}                                                                                                              & \textbf{Author(s)}                     & \textbf{\begin{tabular}[c]{@{}l@{}}Number\\ of classes\end{tabular}} & \textbf{\begin{tabular}[c]{@{}l@{}}Text length\\ (25, 75)\\ percentiles\end{tabular}} \\ \hline
\href{https://huggingface.co/datasets/ag_news}{\texttt{ag\_news}}                                                                          & \citet{zhang2015character}             & 4                                                                    & (196, 266)                                                                            \\ \hline
\href{https://huggingface.co/datasets/SetFit/amazon_counterfactual_en}{\texttt{SetFit/amazon\_counterfactual\_en}}                         & \citet{oneill-etal-2021-wish}          & 2                                                                    & (60, 125)                                                                             \\ \hline
\href{https://huggingface.co/datasets/app_reviews}{\texttt{app\_reviews}}                                                                  & \citet{grano2017android}               & 5                                                                    & (10, 77)                                                                              \\ \hline
\href{https://huggingface.co/datasets/blog_authorship_corpus}{\texttt{blog\_authorship\_corpus}}                                           & \citet{schler2006effects}              & 2                                                                    & (92, 556)                                                                             \\ \hline
\href{https://huggingface.co/datasets/christinacdl/clickbait_notclickbait_dataset}{\texttt{christinacdl/clickbait\_notclickbait\_dataset}} &                                        & 2                                                                    & (46, 69)                                                                              \\ \hline
\href{https://huggingface.co/datasets/climate_fever}{\texttt{climate\_fever}}                                                              & \citet{diggelmann2020climatefever}     & 4                                                                    & (80, 156)                                                                             \\ \hline
\href{https://huggingface.co/datasets/aladar/craigslist_bargains}{\texttt{aladar/craigslist\_bargains}}                                    & \citet{he-etal-2018-decoupling}        & 6                                                                    & (346, 713)                                                                            \\ \hline
\href{https://huggingface.co/datasets/disaster_response_messages}{\texttt{disaster\_response\_messages}}                                   &                                        & 3                                                                    & (74, 178)                                                                             \\ \hline
\href{https://huggingface.co/datasets/emo}{\texttt{emo}}                                                                                   & \citet{chatterjee-etal-2019-semeval}   & 4                                                                    & (44, 83)                                                                              \\ \hline
\href{https://huggingface.co/datasets/dair-ai/emotion}{\texttt{dair-ai/emotion}}                                                           & \citet{saravia-etal-2018-carer}        & 6                                                                    & (53, 129)                                                                             \\ \hline
\href{https://huggingface.co/datasets/SetFit/enron_spam}{\texttt{SetFit/enron\_spam}}                                                      & \citet{metsis2006spam}                 & 2                                                                    & (342, 1553)                                                                           \\ \hline
\href{https://huggingface.co/datasets/financial_phrasebank}{\texttt{financial\_phrasebank}}                                                & \citet{Malo2014GoodDO}                 & 3                                                                    & (79, 157)                                                                             \\ \hline
\href{https://huggingface.co/datasets/classla/FRENK-hate-en}{\texttt{classla/FRENK-hate-en}}                                               & \citet{ljubešić2019frenk}              & 2                                                                    & (34, 160)                                                                             \\ \hline
\href{https://huggingface.co/datasets/hyperpartisan_news_detection}{\texttt{hyperpartisan\_news\_detection}}                               & \citet{kiesel-etal-2019-semeval}       & 2                                                                    & (39, 63)                                                                              \\ \hline
\href{https://huggingface.co/datasets/limit}{\texttt{limit}}                                                                               & \citet{manotas-etal-2020-limit}        & 2                                                                    & (53, 123)                                                                             \\ \hline
\href{https://huggingface.co/datasets/AmazonScience/massive}{\texttt{AmazonScience/massive}}                                               & \citet{fitzgerald-etal-2023-massive}   & 18                                                                   & (24, 44)                                                                              \\ \hline
\href{https://huggingface.co/datasets/movie_rationales}{\texttt{movie\_rationales}}                                                        & \citet{deyoung-etal-2020-eraser}       & 2                                                                    & (2721, 4659)                                                                          \\ \hline
\href{https://huggingface.co/datasets/mteb/mtop_domain}{\texttt{mteb/mtop\_domain}}                                                        & \citet{muennighoff-etal-2023-mteb}     & 11                                                                   & (26, 44)                                                                              \\ \hline
\href{https://huggingface.co/datasets/ccdv/patent-classification}{\texttt{ccdv/patent-classification}}                                     & \citet{sharma-etal-2019-bigpatent}     & 9                                                                    & (441, 775)                                                                            \\ \hline
\href{https://huggingface.co/datasets/rotten_tomatoes}{\texttt{rotten\_tomatoes}}                                                          & \citet{pang-lee-2005-seeing}           & 2                                                                    & (76, 149)                                                                             \\ \hline
\href{https://huggingface.co/datasets/silicone}{\texttt{silicone}}                                                                         & \citet{chapuis-etal-2020-hierarchical} & 4                                                                    & (29, 75)                                                                              \\ \hline
\href{https://huggingface.co/datasets/trec}{\texttt{trec}}                                                                                 & \citet{wang-etal-2007-jeopardy}        & 6                                                                    & (36, 61)                                                                              \\ \hline
\href{https://huggingface.co/datasets/tweets_hate_speech_detection}{\texttt{tweets\_hate\_speech\_detection}}                              & \citet{sharma2019}                     & 2                                                                    & (62, 107)                                                                             \\ \hline
\href{https://huggingface.co/datasets/yahoo_answers_topics}{\texttt{yahoo\_answers\_topics}}                                               & \citet{zhang2018}                      & 10                                                                   & (58, 213)                                                                             \\ \hline
\href{https://huggingface.co/datasets/yelp_review_full}{\texttt{yelp\_review\_full}}                                                       & \citet{zhang2015character}             & 5                                                                    & (287, 957)                                                                            \\ \hline
\end{tabular}
\caption{Brief descriptions of the 25 classification tasks used in this experiment. Click the link in the cell to be taken to the dataset homepage in \url{https://huggingface.co/datasets}. The dataset subset (or config) and the chosen prediction task are specified in code in \href{https://github.com/kddubey/pretrain-on-test/blob/main/src/pretrain_on_test/data.py}{\texttt{src/pretrain\_on\_test/data.py}}.}
\label{tab:tasks}
\end{table*}

The experiment was ran on 25 publicly available text classification tasks found in \url{https://huggingface.co/datasets}.
Inclusion criteria:

\begin{enumerate}[itemsep=0em]
    \item All text is in English.
    \item The number of classes is not greater than 25, because only 50 or 100 observations are used for training the classifier.
    \item The task is to classify one text, not a pair as in, e.g., textual entailment tasks.
    \item Texts are not so long that too much useful signal is dropped when text is truncated to fit in BERT/GPT-2's context window, which is set to 256 tokens.
    \item Based on our best judgment, it is likely that BERT/GPT-2 can do better than guessing.
\end{enumerate}

Table~\ref{tab:tasks} lists the exact tasks.

\section{Other experiment choices}
\label{sec:choices}

This section expands on \S \ref{sec:exp}.

First, we clarify how classification training is performed. For BERT, the linear layer transforms the \texttt{[CLS]} token embedding. For GPT-2, the linear layer transforms the last token's embedding. The output dimension of the linear layer is the number of classes in the classification task. This layer, along with the rest of the weights in the LM, are finetuned to minimize classification cross entropy loss on \texttt{train}.

The BERT model used here is \href{https://huggingface.co/google-bert/bert-base-uncased}{\texttt{bert-base-uncased}}. The GPT-2 model used here is \href{https://huggingface.co/openai-community/gpt2}{\texttt{gpt2}} (small), with 124M parameters.

\texttt{train} is stratify-sampled by the class to ensure every class is represented, and to reduce the variance of accuracy estimators. \texttt{test} is not stratify-sampled. We are only interested in the \textit{difference} between accuracies, which is a function of the difference between model likelihoods because the priors are uniform. So even if accuracies are worse than the majority vote, differences are still meaningful for the purposes of this experiment.

\texttt{train} text is not included during pretraining to eliminate the overlap of pretraining data between acc\textsubscript{extra} and acc\textsubscript{test}. This choice was made in an effort to widen any gap between them. 
% The experiment tries to go out of its way to provide evidence of a bias.

\texttt{train} contains $m = 50$ or $m = 100$ observations. $m = 50$ is inspired by the RAFT benchmark. $m=100$ stretches the intention of "few" in few-shot learning, but was tested in an attempt to make lower-variance comparisons. BERT is quite sensitive---see Appendix~\ref{sec:diff_distr}.

% The discussion focuses on the BERT and GPT-2 results because their pretraining data is (likely) not already contaminated with text from the 25 text classification tasks. While modern NLP solutions usually involve LLMs, these models' pretraining data are opaque and more likely to include text from the 25 classification tasks (for example, from web-crawling the Dataset Viewer in HuggingFace's datasets web pages, which hosts the experiment's data) \citep{jacovi-etal-2023-stop}. As a result, the comparisons---acc\textsubscript{extra} versus acc\textsubscript{base} and acc\textsubscript{test} versus acc\textsubscript{extra}---would  be less valid.

\section{Hyperparameters and reproducibility}
\label{sec:hps}

This paper's experiment and analysis code, and data, is available here: \url{https://github.com/kddubey/pretrain-on-test}.

\href{https://github.com/kddubey/pretrain-on-test/blob/main/experiment.sh}{\texttt{experiment.sh}} lists hyperparameters used for each classification task and experiment configuration. For the experiment in \S \ref{sec:exp}, BERT was pretrained for 2 epochs, and GPT-2 was pretrained for 1 epoch. Classification hyperparameters were pre-specified based on \citet{zhang2021revisiting}, with batch sizes set to avoid out-of-memory errors. Run the script on a GPU with at least 15 GB RAM to reproduce results in \S \ref{sec:results}. It takes about 5 days on a T4 GPU. Training is performed using the \texttt{transformers} package \citep{wolf-etal-2020-transformers}.

\section{Results}

\subsection{Task-level analysis}

The notebook \href{https://github.com/kddubey/pretrain-on-test/blob/main/analysis/dataset.ipynb}{\texttt{analysis/dataset.ipynb}} can be run to (1) produce visualizations of the distributions of acc\textsubscript{extra}, acc\textsubscript{test}, and acc\textsubscript{base} (for each classification task and experiment configuration), and (2) compute $p$-values for the hypothesis test specified in \eqref{eq:null}. For all settings of $m$ and $n$, no $p$-values were statistically significant at the $0.05$ level.
% \footnote{For the test that $\text{E}[\textcolor{blue}{\text{acc}_{\text{extra}} - \text{acc}_{\text{base}}}] < 0$, 16 of the 25 $p$-values for $m = 50, n = 50$ in \S \ref{sec:overtraining} were less than $0.05$, which demonstrates that the test has some statistical power.}

In Figure~\ref{fig:lemons_bert}, \texttt{amazon\_counterfactual\_en} and \texttt{mtop\_domain} have a consistent evaluation bias across $m$ for $n = 500$ and $n = 200$, respectively. But these tasks did not result in an evaluation bias in any other experiment configuration, including those with GPT-2 and Mistral 7B.

Care has to be taken when attempting to analyze or interpret \textcolor{blue}{acc\textsubscript{extra} $-$ acc\textsubscript{base}} and \textcolor{orange}{acc\textsubscript{test} $-$ acc\textsubscript{extra}} \textit{together}. That's because these differences are not independent: if acc\textsubscript{extra} is high, then \textcolor{blue}{acc\textsubscript{extra} $-$ acc\textsubscript{base}} increases and \textcolor{orange}{acc\textsubscript{test} $-$ acc\textsubscript{extra}} decreases. This paper does not analyze the scores together, per se. We care about \textcolor{orange}{acc\textsubscript{test} $-$ acc\textsubscript{extra}}. \textcolor{blue}{acc\textsubscript{extra} $-$ acc\textsubscript{base}} only exists to sanity check that the pretraining code works; there may be an effect to detect.

\subsection{Difference distributions}
\label{sec:diff_distr}

\begin{figure*}
    \centering
    \includegraphics[height=0.8\paperheight]{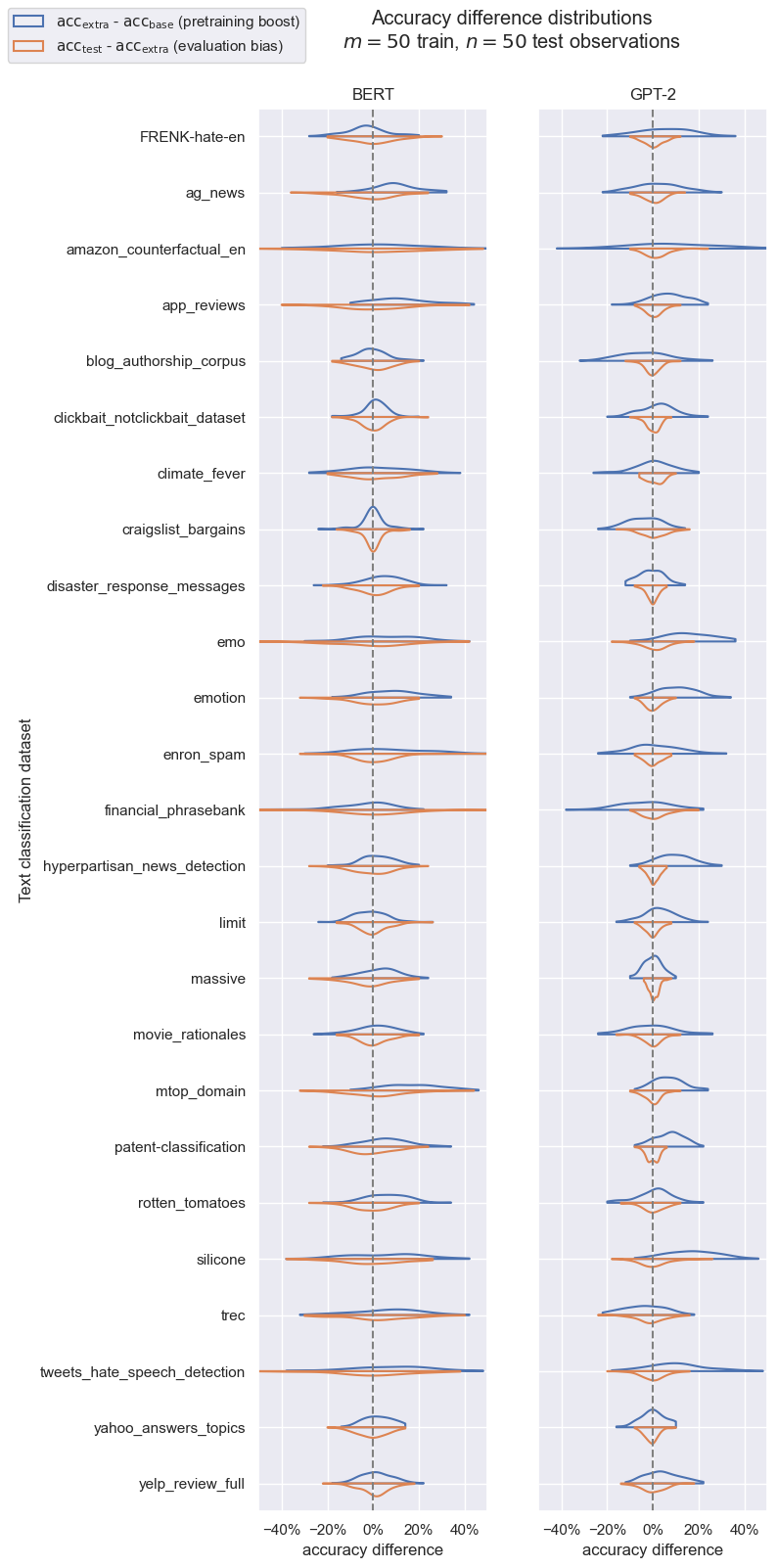}
    \caption{}
    \label{fig:diff_distr_m50_n50}
\end{figure*}

\begin{figure*}
    \centering
    \includegraphics[height=0.8\paperheight]{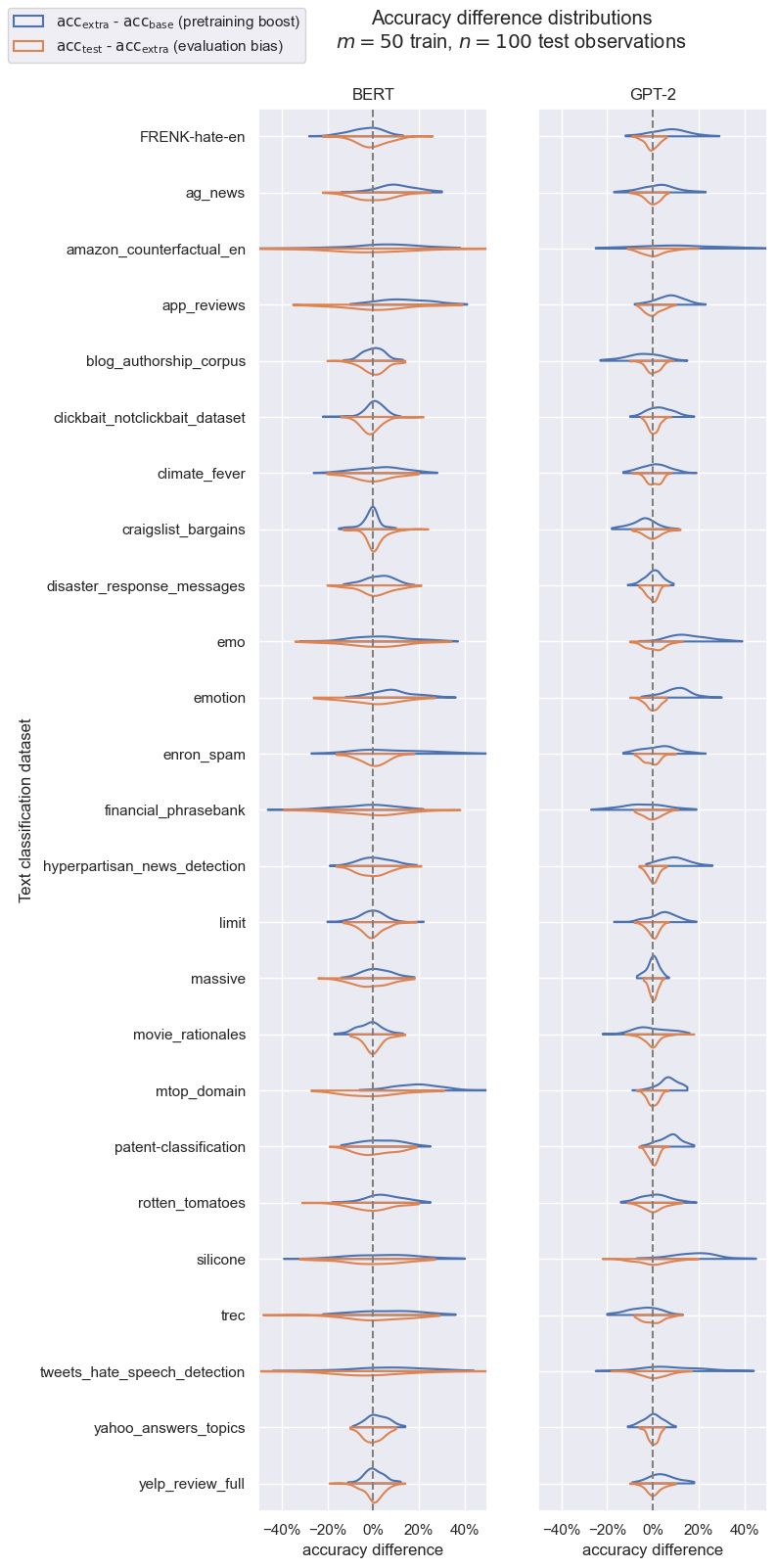}
    \caption{}
    \label{fig:diff_distr_m50_n100}
\end{figure*}

\begin{figure*}
    \centering
    \includegraphics[height=0.8\paperheight]{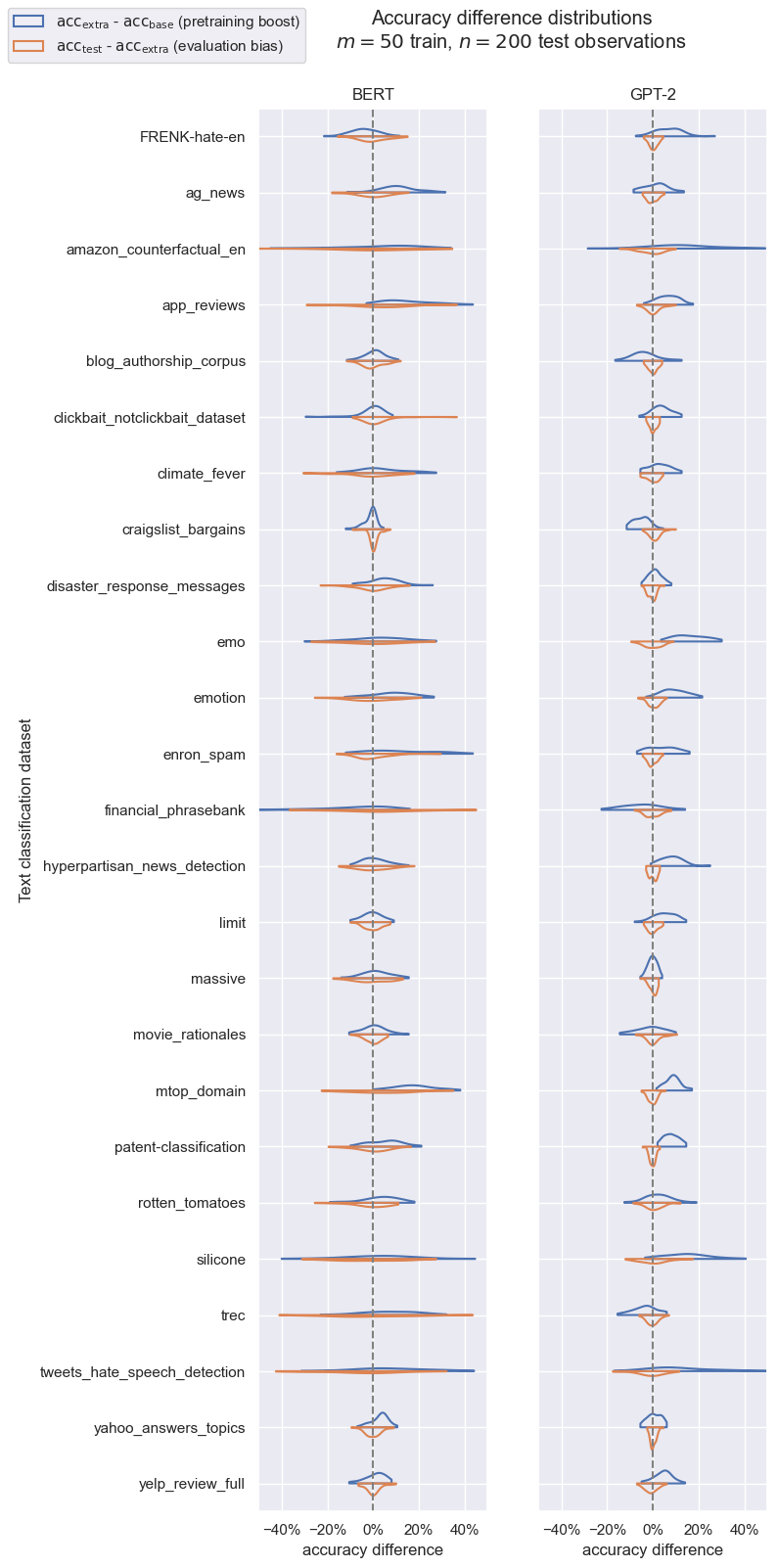}
    \caption{}
    \label{fig:diff_distr_m50_n200}
\end{figure*}

\begin{figure*}
    \centering
    \includegraphics[height=0.8\paperheight]{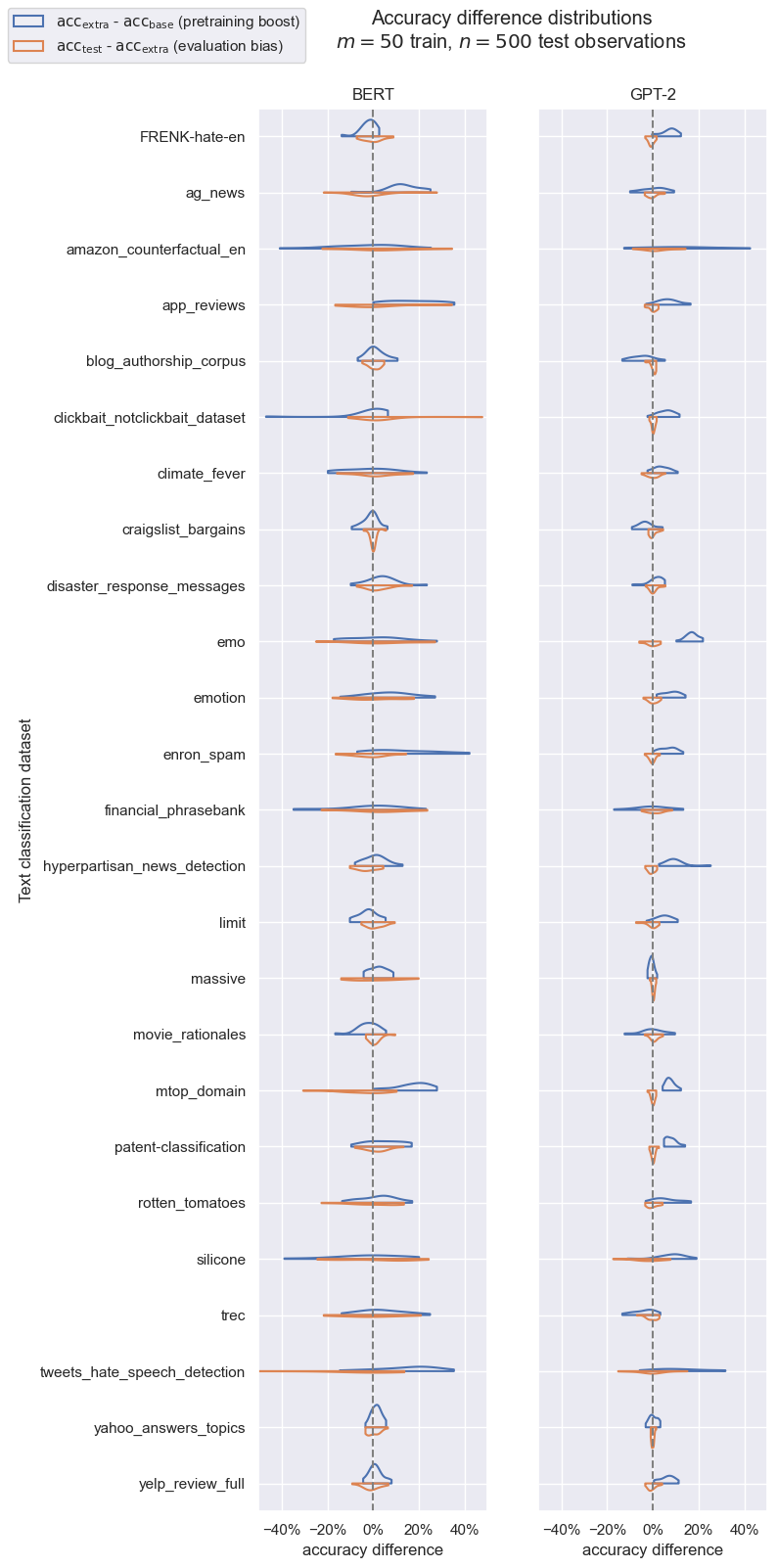}
    \caption{}
    \label{fig:diff_distr_m50_n500}
\end{figure*}

\begin{figure*}
    \centering
    \includegraphics[height=0.8\paperheight]{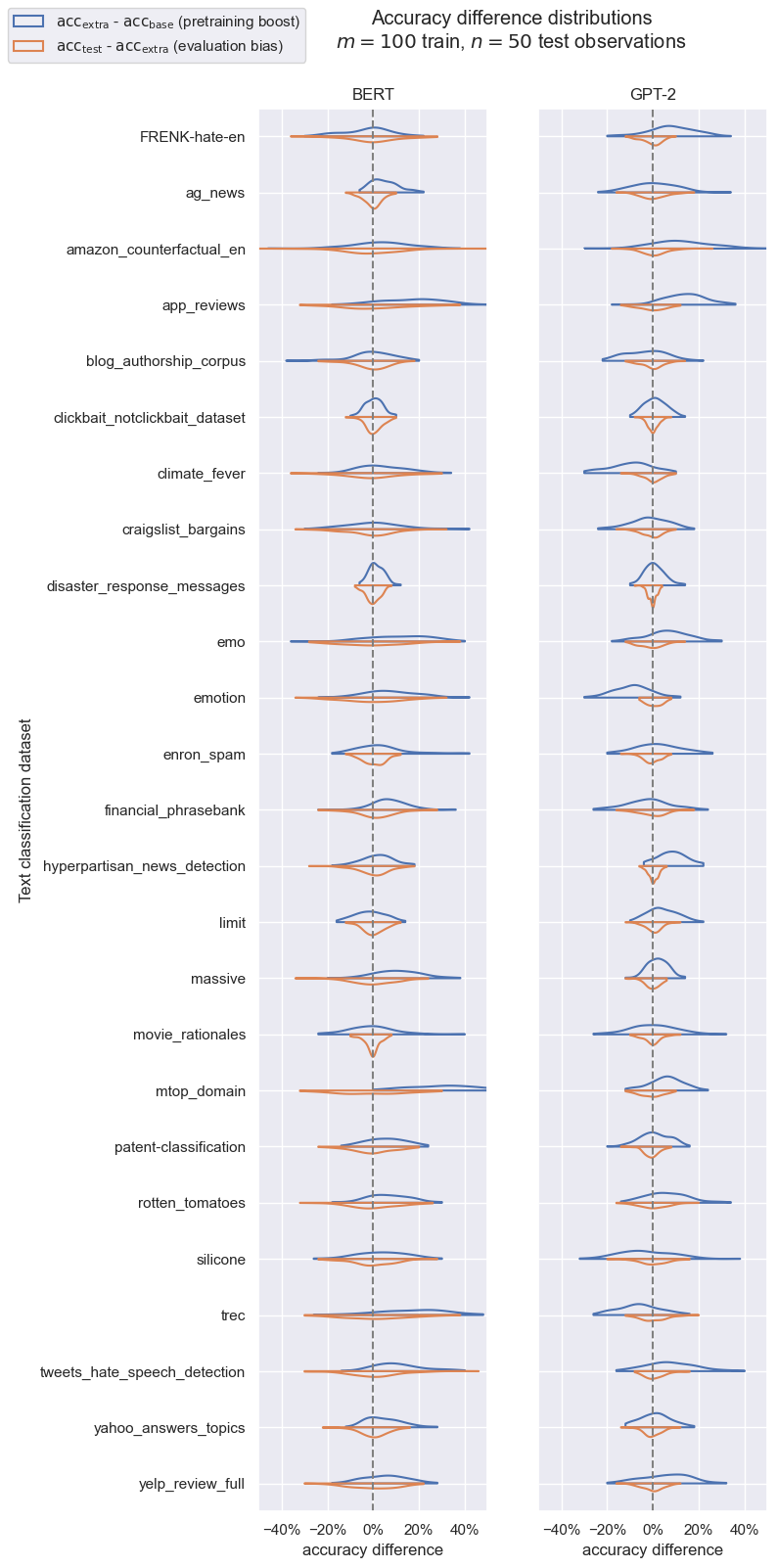}
    \caption{}
    \label{fig:diff_distr_m100_n50}
\end{figure*}

\begin{figure*}
    \centering
    \includegraphics[height=0.8\paperheight]{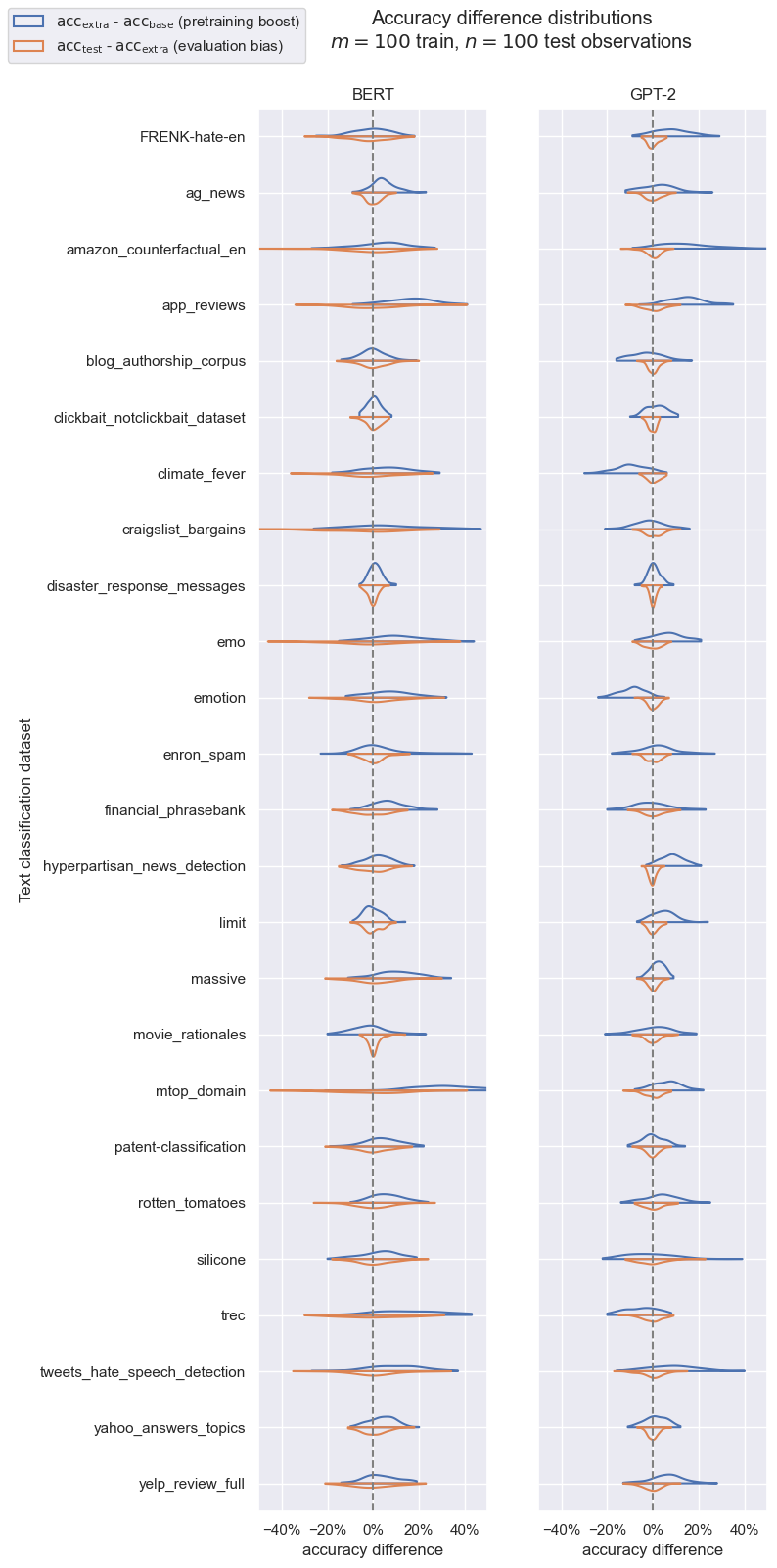}
    \caption{}
    \label{fig:diff_distr_m100_n100}
\end{figure*}

\begin{figure*}
    \centering
    \includegraphics[height=0.8\paperheight]{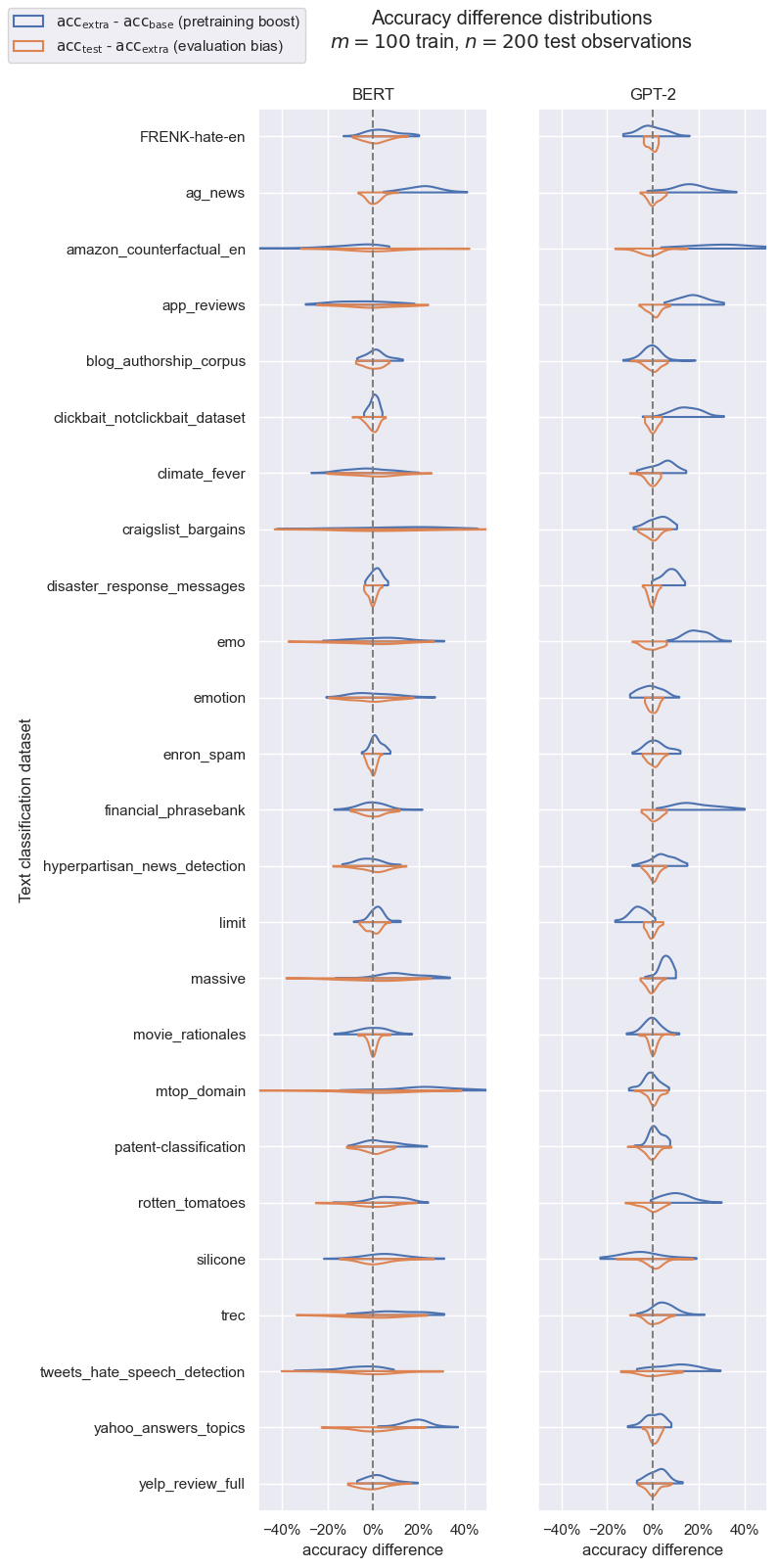}
    \caption{}
    \label{fig:diff_distr_m100_n200}
\end{figure*}

\begin{figure*}
    \centering
    \includegraphics[height=0.8\paperheight]{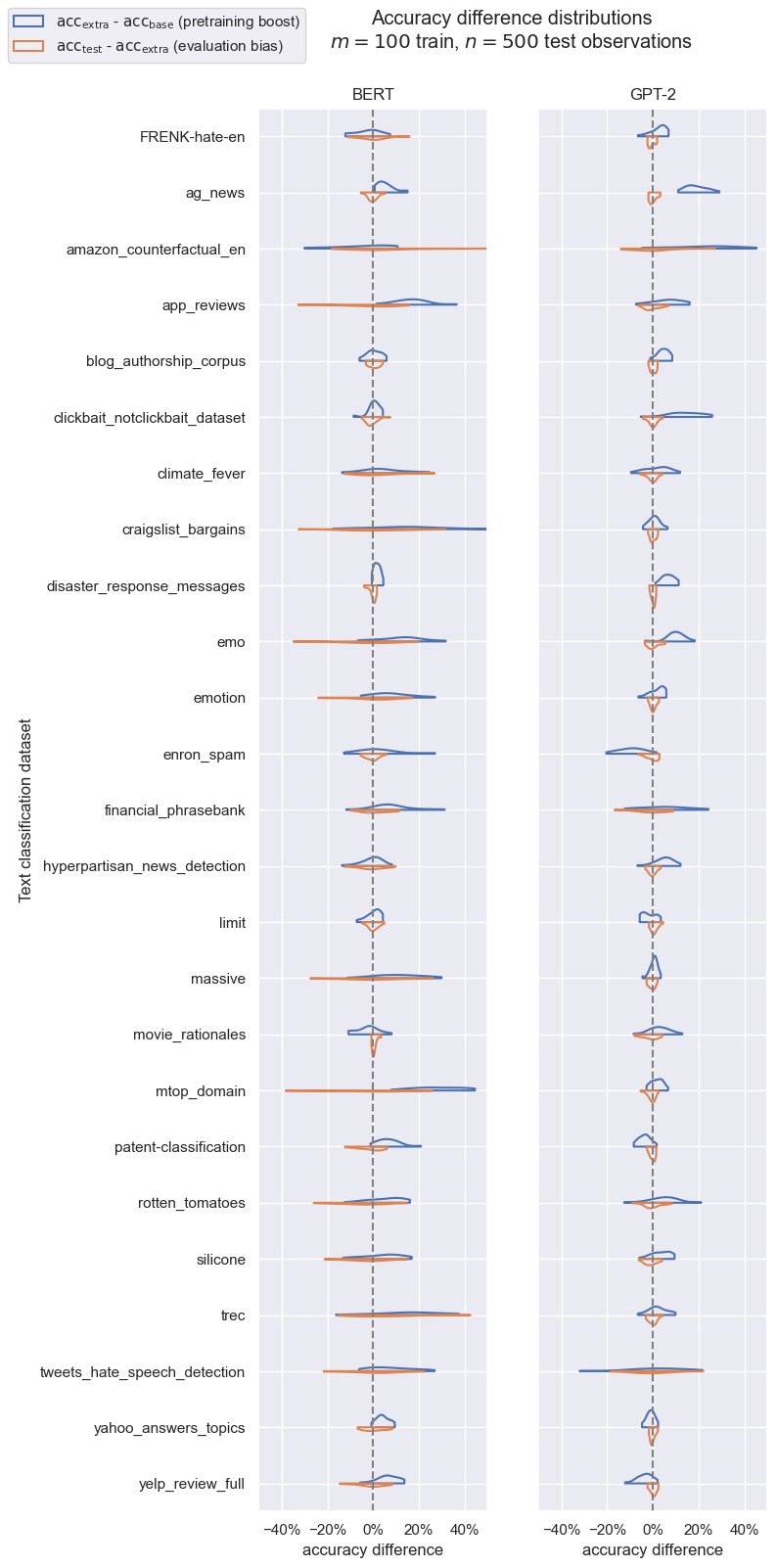}
    \caption{}
    \label{fig:diff_distr_m100_n500}
\end{figure*}

Figures \ref{fig:diff_distr_m50_n50} - \ref{fig:diff_distr_m100_n500} visualize the distributions of the paired differences---\textcolor{blue}{acc\textsubscript{extra} $-$ acc\textsubscript{base}} and \textcolor{orange}{acc\textsubscript{test} $-$ acc\textsubscript{extra}}--for each configuration of the experiment.

\section{Analysis}
\label{sec:analysis-appendix}

The analysis in \S \ref{sec:analysis} can be reproduced by running all of the notebooks in \href{https://github.com/kddubey/pretrain-on-test/tree/main/analysis/fit_posteriors}{\texttt{analysis/fit\_posteriors/}}. Figure~\ref{fig:main} can be reproduced by running the notebook \href{https://github.com/kddubey/pretrain-on-test/blob/main/analysis/results/posterior_pred.ipynb}{\texttt{analysis/results/posterior\_pred.ipynb}}. Figure~\ref{fig:lemons_bert} can be reproduced by running the notebook \href{https://github.com/kddubey/pretrain-on-test/blob/main/analysis/results/posterior_pred_conditional.ipynb}{\texttt{analysis/\\
results/posterior\_pred\_conditional.ipynb}}. Changing the threshold for the bias to +2\% accuracy instead of +3\% did not change conclusions.

Posterior samples of $\beta$ (which were used to draw posterior predictive samples) were taken from four chains with 1,000 draws each, after 500 steps of tuning. 

\subsection{Hierarchical model checks}
\label{sec:model_checks}

Hierarchical models require some basic checks to have faith in their results \citep{mcelreath2018statistical}.

For each of the 24 hierarchical models (16 in \S \ref{sec:discussion}, 4 in \S \ref{sec:overtraining}, and 4 in \S \ref{sec:zero_shot}), no divergences were observed during the fitting procedure. All trace plots were healthy.

\begin{figure*}
  \begin{subfigure}{{0.5\textwidth}}
    % \centering
    \includegraphics[width=0.9\linewidth]{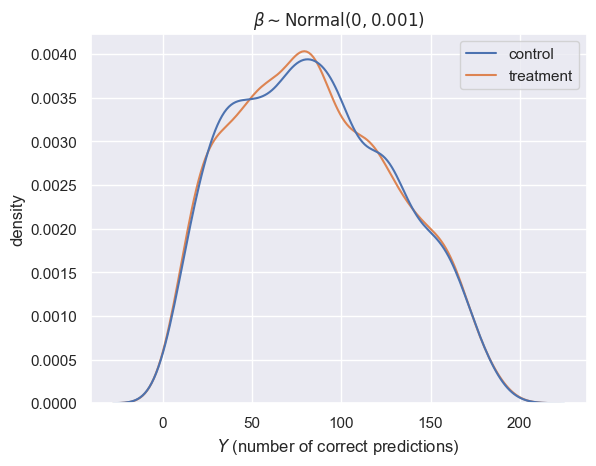}
    \caption{Null effect}
  \end{subfigure}%
  \begin{subfigure}{{0.5\textwidth}}
    % \centering
    \includegraphics[width=0.9\linewidth]{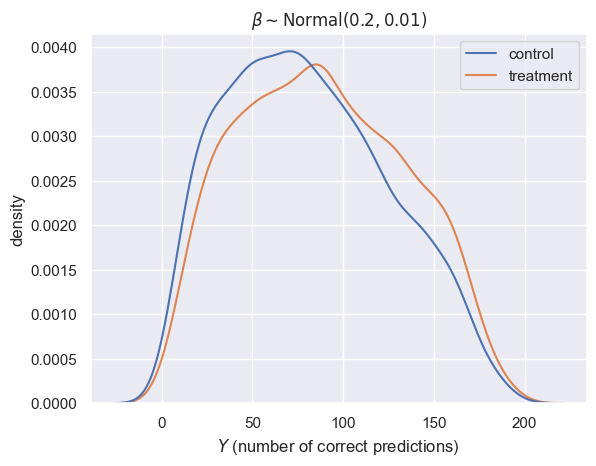}
    \caption{Non-null, positive effect}
  \end{subfigure}

  \caption{Prior predictive distributions for $m=100, n=200$ from two different priors for $\beta$---the expected increase in the log-odds of a correct prediction resulting from an intervention/treatment.}
  \label{fig:prior_pred}
\end{figure*}

Figure~\ref{fig:prior_pred} contains prior predictive distributions for $m=100, n=200$, demonstrating that priors are not unreasonable. Using default priors from the \texttt{bambi} package \citep{Capretto2022}, while scientifically unreasonable (because they result in wide, basin-like accuracy distributions), did not change the conclusions of this paper.

\begin{figure*}

  \begin{subfigure}{\textwidth}
    \centering
    \includegraphics[width=\linewidth]{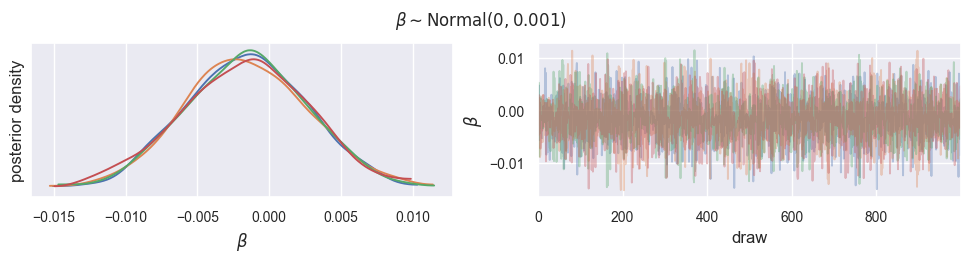}
    \caption{Null effect is recovered.}
  \end{subfigure} \\
  \begin{subfigure}{\textwidth}
    \centering
    \includegraphics[width=\linewidth]{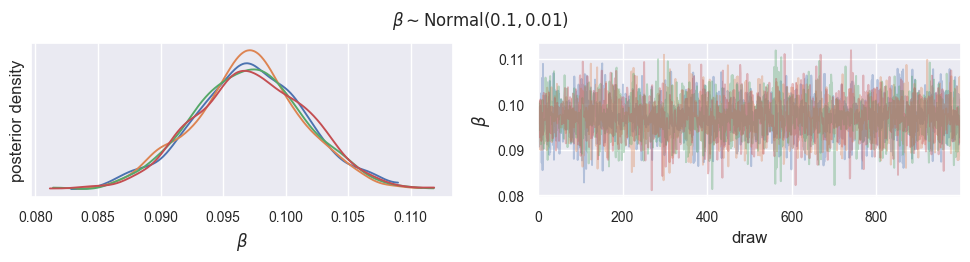}
    \caption{Non-null, positive effect is recovered.}
  \end{subfigure}

  \caption{Posterior distributions and trace plots for null and non-null effects \textbf{from simulated data} where $m=100, n=200$, approximated by four chains with 1,000 draws each, after 500 steps of tuning. For each model, no divergences were observed during the fitting procedure. Visualizations were produced by the \texttt{arviz} package \citep{arviz_2019}.}
  \label{fig:test_model}
\end{figure*}

Figure~\ref{fig:test_model} contains posterior distributions of $\beta$ for $m=100, n=200$, demonstrating the hierarchical model's ability to recover both null and non-null effects. This test can be reproduced by running the notebook \href{https://github.com/kddubey/pretrain-on-test/blob/main/analysis/test.ipynb}{\texttt{analysis/test.ipynb}}.

\begin{figure*}
\centering
\includegraphics[width=0.9\linewidth]{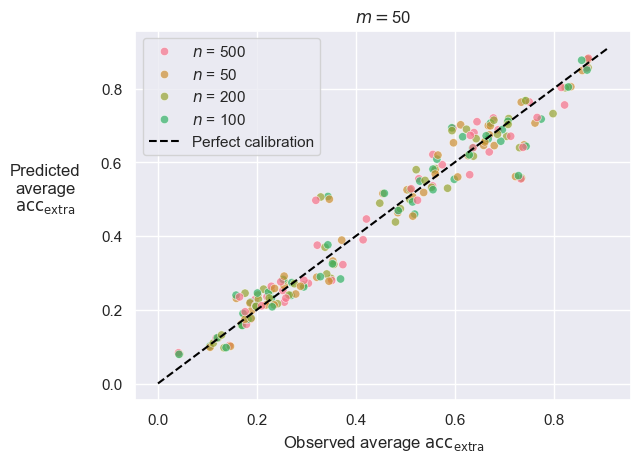} \hfill
\caption{Each of the points represents a task and an LM type (BERT or GPT-2).}
\label{fig:task_calplot}
\end{figure*}

\begin{figure*}
\includegraphics[width=0.48\linewidth]{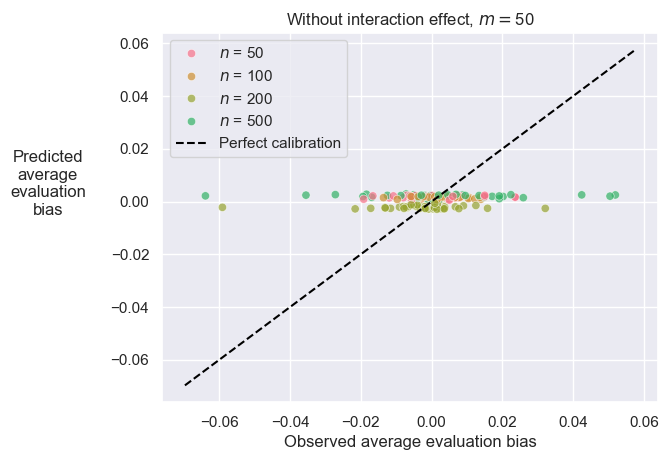}
\includegraphics[width=0.48\linewidth]{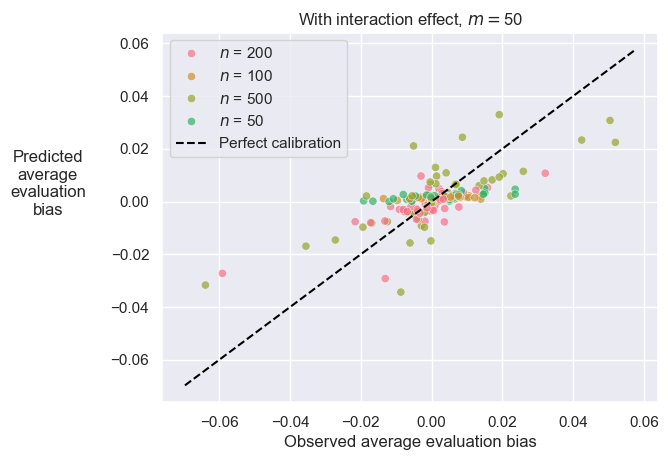} \hfill
\caption{Omitting the interaction effect causes underfitting. Note that the prior causes effects to shrink towards 0. Each of the points represents a task and an LM type (BERT or GPT-2).}
\label{fig:interaction}
\end{figure*}

Figure~\ref{fig:task_calplot} checks that posterior predictions for the average task accuracies are calibrated. Figure~\ref{fig:interaction} demonstrates the importance of including the $W_{jl}$ term. These figures can be reproduced by running the notebook \href{https://github.com/kddubey/pretrain-on-test/blob/main/analysis/results/posterior_pred_conditional.ipynb}{\texttt{analysis/results/\\
posterior\_pred\_conditional.ipynb}}.

\section{Meta-analysis}
\label{sec:meta-appendix}

The meta-analysis in \S \ref{sec:meta} can be reproduced by running the script, \href{https://github.com/kddubey/pretrain-on-test/blob/main/analysis/meta/meta.py}{\texttt{analysis/meta/meta.py}}, and then the notebook \href{https://github.com/kddubey/pretrain-on-test/blob/main/analysis/meta/meta.ipynb}{\texttt{analysis/meta/meta.ipynb}}. No divergences were observed.

Another question is whether the subsample causes a consistent evaluation bias. \S \ref{sec:meta} establishes that picking a single subsample causes the comparison between acc\textsubscript{test} and acc\textsubscript{extra} to be a coin flip. But is the result of the coin flip explained by the specific subsample that was drawn? If so, comparing models using a single subsample may not be so noisy, because the effect of pretraining on unlabeled test set text would be consistent across models.

One way to answer this question is to measure the correlation between the evaluation bias of BERT and GPT-2 for each setting of $m$ and $n$, and each of the 25 tasks. A positive correlation suggests that the subsample causes the evaluation bias. 
Spearman's rank correlation coefficient is used because we are only interested in the consistency of the relationship, not its linearity.

\begin{figure}[htbp]
  \begin{subfigure}{0.85\columnwidth}
    {\centering\includegraphics[width=\linewidth]{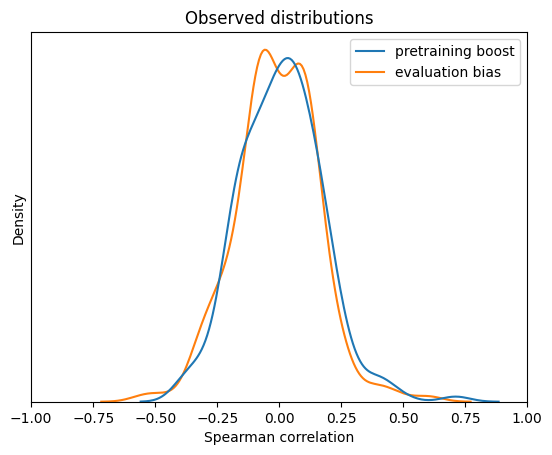}\par}
    \caption{Observed}
  \end{subfigure}
  \hfill
  \begin{subfigure}{0.85\columnwidth}
    \centering
    \includegraphics[width=\linewidth]{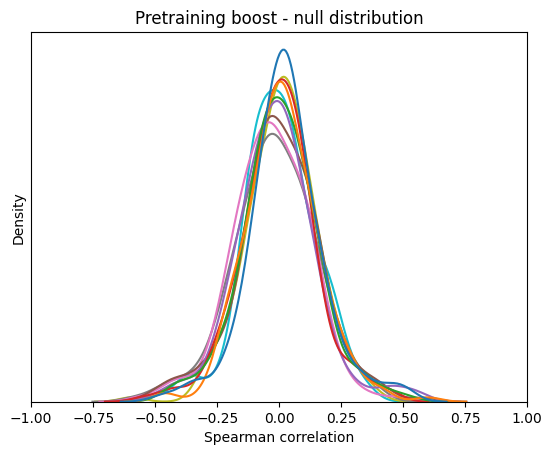}
    \par
    \caption{Randomly permuted}
  \end{subfigure}
  \hfill
  \begin{subfigure}{0.85\columnwidth}
    \centering
    \includegraphics[width=\linewidth]{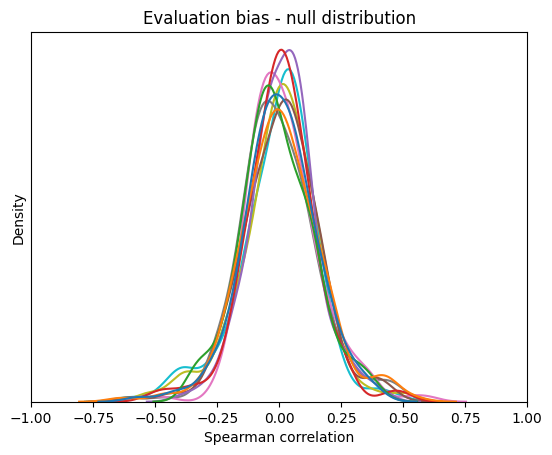}
    \par
    \caption{Randomly permuted}
  \end{subfigure}

  \caption{Distribution of correlation between BERT and GPT-2 across all $m$, $n$, and the 25 classification tasks.}
  \label{fig:meta_corr}
\end{figure}

The observed distributions of correlations across $m$, $n$, and the tasks are plotted in Figure~\ref{fig:meta_corr} (a). For context, 10 distributions of randomly permuted pairs of subsample-level biases are plotted in Figure~\ref{fig:meta_corr} (b) and (c). These correlations are theoretically 0, and are positive or negative by chance alone. The observed distributions are qualitatively indistinguishable from the null ones. Notably, the variance is consistent. A deeper dive into the correlations did not find any consistently positive (or negative) correlations at the task level. This result further evidences the importance of repeated subsampling. Taking a single subsample does not result in a consistent pretraining boost or evaluation bias between BERT and GPT-2. This analysis can be reproduced by running the notebook \href{https://github.com/kddubey/pretrain-on-test/blob/main/analysis/dataset_level.ipynb}{\texttt{analysis/dataset\_level.ipynb}}.

\section{Zero-shot text classification}
\label{sec:zero_shot-appendix}

Here is an example of a prompt for the \href{https://huggingface.co/datasets/fancyzhx/ag_news}{\texttt{ag\_news}} task \citep{zhang2015character}:

\small
\begin{verbatim}
    Your task is to classify a given text as
    one of these categories:
    World
    Sports
    Business
    Sci/Tech
    
    The text is a news article. Answer with its topic.
    
    ### Text: Bombardier CEO Quits, Shares
    Dive Paul Tellier stepped down on Monday
    as president and chief executive of
    Bombardier Inc. (BBDsvb.TO: Quote,
    Profile, Research) (BBDb.
     ### Answer:
\end{verbatim}
\normalsize

For packing (\S \ref{sec:packing}), the prompts at inference are in the same format as above. For training, 8 texts were packed. Here is an example of an input sequence for \href{https://huggingface.co/datasets/fancyzhx/ag_news}{\texttt{ag\_news}}, where 4 texts are packed:

\small
\begin{verbatim}
    Your task is to classify a given text as
    one of these categories:
    World
    Sports
    Business
    Sci/Tech
    
    The text is a news article. Answer with its topic.
    
    ### Text: US Electoral College withstands critics
    ... so far (AFP) AFP - Lambasted as antiquated 
    and anti-democratic, the Electoral College that 
    decides the US presidency has survived for 
    centuries as an unmovable albeit creaky pillar 
    of the American political system.
    
    
    ### Text: Voters in Hungary decide referenda 
    Voters in Hungary went to the polls Sunday to 
    decide a double referendum on citizenship 
    rights and their nation #39;s health care 
    system.
    
    
    ### Text: White House: Trying to Confirm Terror 
    Group #39;s Allegiance to bin 
    &lt;b&gt;...&lt;/b&gt; The Bush administration 
    says it #39;s trying to confirm the latest 
    declaration from the most feared militant group 
    in Iraq. In a statement posted on a Web site 
    Sunday, the group led by terror mastermind Abu 
    Musab
    
    
    ### Text: Fans rush to create mods for 
    long-awaited  #39;Doom 3 #39; Activision #39;s 
    Doom 3, which launched earlier this month, 
    wasn #39;t on store shelves for three days 
    before players started creating their own
    modifications - known as mods -o the game.
\end{verbatim}
\normalsize

The zero-shot experiment files are in \href{https://github.com/kddubey/pretrain-on-test/tree/main/cloud_scripts/gcp/experiments/zero-shot}{\texttt{cloud\_scripts/gcp/experiments/zero\_shot/}} and \href{https://github.com/kddubey/pretrain-on-test/tree/main/cloud_scripts/gcp/experiments/zero-shot-packing}{\texttt{cloud\_scripts/gcp/experiments/\\zero\_shot\_packing/}}. Batch sizes are set to run on a GPU with at least 20 GB RAM. The GPU must support the data types needed for QLoRA, e.g., an L4 GPU. Figure~\ref{fig:zero_shot} can be reproduced by running the notebooks in \href{https://github.com/kddubey/pretrain-on-test/tree/main/analysis/fit_posteriors/zero_shot}{\texttt{analysis/fit\_posteriors/zero\_shot}} and \href{https://github.com/kddubey/pretrain-on-test/tree/main/analysis/fit_posteriors/zero_shot_packing}{\texttt{analysis/fit\_posteriors/zero\_shot\_packing}} and then the notebook, \href{https://github.com/kddubey/pretrain-on-test/blob/main/analysis/results/posterior_pred.ipynb}{\texttt{analysis/results/posterior\_pred.ipynb}}.

The Mistral 7B model is \href{https://huggingface.co/mistralai/Mistral-7B-v0.3}{Mistral-7B-v0.3}, the non-instruction-trained model.

We only study $n=100$ in an initial effort to provide evidence of an evaluation bias (due to the relatively small test set), and take 20 repeated subsamples instead of 50. While $n=100$ is quite small, benchmarks such as LegalBench \citep{guha2024legalbench} have test data in this range. And the analysis transparently exposes variance.

QLoRA hyperparameters were pre-specified: every adapter has rank 16 with $\alpha = 32$ (LoRA scaling factor), a $0.05$ dropout rate, and no bias parameters. The adapter layers introduce $41,943,040$ new, trainable parameters to Mistral 7B, whose parameters are frozen. Pretraining was done for 1 epoch.

% Classification is performed by picking the class name with the highest average token log-probability (conditional on the prompt), as done in \citet{trinh2018simple}. For tasks where class names are each 1 token long, this classification strategy is equivalent to picking the highest-probability answer.

To increase the power of the contamination hypothesis test run in \S \ref{sec:contamination}, shards were formed to be similar to the sequences passed in during pretraining. Here is an example of what the first 2 text-label pairs in the dataset passed to the contamination test looks like:

\small
\begin{verbatim}

    ### Text: Customers bemoan changes in Quicken 
    2005 The new version of the personal finance 
    program drops support for a widely used file 
    format.
     ### Answer: Sci/Tech

    ### Text: Blair gives partial Iraq apology Tony 
    Blair has offered his Labour party a partial 
    apology for waging war in Iraq, striving to pull 
    angry supporters behind him ahead of an election 
    next year.
     ### Answer: World

\end{verbatim}
\normalsize

The 2 $p$-values in \S \ref{sec:contamination} can be obtained by running the notebook \href{https://github.com/kddubey/pretrain-on-test/blob/main/analysis/contamination/test.ipynb}{\texttt{analysis/contamination/test.ipynb}} on an L4 GPU.

% \section{Experiment with PCA}
% \label{sec:pca}

% This is a tangential experiment which served as part of the inspiration of the paper.

% This experiment's design is akin to \S \ref{sec:exp} except that the unsupervised pretraining procedure is PCA, the supervised training procedure is linear regression on synthetic data, and $2,000$ subsamples are taken instead of 20-100. 20 features are generated with known effective rank. The dependent variable is a linear transformation of these features plus independent, normally distributed noise with a standard deviation of 1. Performance is measured using $R^2$.

% Figure~\ref{fig:pca} shows that the evaluation bias increases with the effective rank of the features. In other words, fitting a PCA on test set features artificially improves test set performance particularly when fitting a PCA on independent features is unhelpful.

% Figure~\ref{fig:pca} can be reproduced by running the notebook \href{https://github.com}{\texttt{analysis/pca.ipynb}}

% \begin{figure*}[!b]
% \centering
% \includegraphics[width=0.8\linewidth]{figures/appendix/pca.png} \hfill
% \caption{The shaded regions are 95\% confidence intervals.}
% \label{fig:pca}
% \end{figure*}

% Dumping big figures last

% m50_n50

% m50_n100

% m50_n200

% m50_n500

% m100_n50

% m100_n100

% m100_n200

% m100_n500

\end{document}